\documentclass[preprint,12pt]{elsarticle}
\usepackage{graphicx}
\usepackage{amssymb}
\usepackage{lineno}
\usepackage{multirow}
\usepackage{amsmath,amssymb}
\usepackage{framed,color}
\definecolor{shadecolor}{rgb}{1,0.8,0.3}
\usepackage[boxruled,vlined,linesnumbered]{algorithm2e}

\journal{PLoS One (Accepted)}
\begin{document}
\begin{frontmatter}


\title{A Hybrid Q-Learning Sine-Cosine-based Strategy for Addressing the Combinatorial Test Suite Minimization Problem}

\author{Kamal Z. Zamli}

\address{IBM Centre of Excellence, Faculty of Computer Systems and Software Engineering, Universiti Malaysia Pahang
Lebuhraya Tun Razak, 26300 Kuantan, Pahang Darul Makmur, Malaysia\\ email: kamalz@ump.edu.my}

\author{Fakhrud Din}

\address{IBM Centre of Excellence, Faculty of Computer Systems and Software Engineering, Universiti Malaysia Pahang
Lebuhraya Tun Razak, 26300 Kuantan, Pahang Darul Makmur, Malaysia \\ email: fakhruddin@uom.edu.pk}

\author{Bestoun S. Ahmed}

\address{Department of Computer Science, Faculty of Electrical Engineering Czech Technical University, Karlovo nam. 13, 121 35 Praha 2, Czech Republic \\ email: albeybes@fel.cvut.cz}

\author{Miroslav Bures}

\address{Department of Computer Science, Faculty of Electrical Engineering Czech Technical University, Karlovo nam. 13, 121 35 Praha 2, Czech Republic \\ email: buresm3@fel.cvut.cz}

\begin{abstract}
The sine-cosine algorithm (SCA) is a new population-based meta-heuristic algorithm. In addition to exploiting sine and cosine functions to perform local and global searches (hence the name sine-cosine), the SCA introduces several random and adaptive parameters to facilitate the search process. Although it shows promising results, the search process of the SCA is vulnerable to local minima/maxima due to the adoption of a fixed switch probability and the bounded magnitude of the sine and cosine functions (from -1 to 1). In this paper, we propose a new hybrid Q-learning sine-cosine- based strategy, called the Q-learning sine-cosine algorithm (QLSCA). Within the QLSCA, we eliminate the switching probability. Instead, we rely on the Q-learning algorithm (based on the penalty and reward mechanism) to dynamically identify the best operation during runtime. Additionally, we integrate two new operations (Lévy flight motion and crossover) into the QLSCA to facilitate jumping out of local minima/maxima and enhance the solution diversity. To assess its performance, we adopt the QLSCA for the combinatorial test suite minimization problem. Experimental results reveal that the QLSCA is statistically superior with regard to test suite size reduction compared to recent state-of-the-art strategies, including the original SCA, the particle swarm test generator (PSTG), adaptive particle swarm optimization (APSO) and the cuckoo search strategy (CS) at the 95\% confidence level. However, concerning the comparison with discrete particle swarm optimization (DPSO), there is no significant difference in performance at the 95\% confidence level. On a positive note, the QLSCA statistically outperforms the DPSO in certain configurations at the 90\% confidence level.
\end{abstract}

\end{frontmatter}

\section{Introduction}
An optimization problem relates to the process of finding the optimal values for the parameters of a given system from all possible values with minimum or maximum profitability. In past decades, many meta-heuristic algorithms have been proposed in the scientific literature (these include genetic algorithms \cite{Ref1}, particle swarm optimization \cite{Ref2}, simulated annealing \cite{Ref3}, and the bat algorithm \cite{Ref4}) to address such a problem. The sine-cosine algorithm (SCA) is a new population-based meta-heuristic algorithm proposed by Mirjalili \cite{Ref5}. In addition to exploiting the sine and cosine functions to perform a local and global search (hence the name sine-cosine), the SCA introduces several random and adaptive parameters to facilitate the search process. Although it shows promising results, the balanced selection of exploration (roaming the random search space on the global scale) and exploitation (exploiting the current good solution in a local region) appears problematic.

Mathematically, sine and cosine are the same operator with a 90-degree phase shift. Therefore, in some cases, the use of either sine or cosine could inadvertently promote similar solutions. Furthermore, the search process is potentially vulnerable to local minima/maxima due to the adoption of a fixed switch probability and the bounded magnitude of the sine and cosine functions (from -1 to 1).

Motivated by these challenges, we propose a new hybrid Q-learning-based sine-cosine strategy called the QLSCA. Hybridization can be the key to further enhancing the performance of the original SCA. Within the QLSCA, we eliminate the switching probability. Instead, we rely on the Q-learning algorithm (based on the penalty and reward mechanism \cite{Ref6}) to dynamically identify the best operation during runtime. Additionally, we combine the QLSCA with two new operations (Lévy flight motion and crossover) to facilitate jumping  out of local minima/maxima and enhance the solution diversity. To assess its performance, we adopt the QLSCA for the combinatorial test suite minimization problem. Experimental results reveal that the QLSCA exhibits competitive performance compared to the original SCA and other meta-heuristic algorithms.

Our contributions can be summarized as follows:

\begin{itemize}

\item A new hybrid Q-learning sine-cosine based strategy called the Q-learning sine-cosine algorithm (QLSCA) that permits the dynamic selection of local and global search operations based on the penalty and reward mechanism within the framework of the Q-learning algorithm.

\item A hybrid of Lévy flight (originated in the cuckoo search algorithm \cite{Ref7}) and crossover (originated in genetic algorithms \cite{Ref1}) operations within the QLSCA.

\item A comparison of the performance of the QLSCA and that of recent state-of-the-art strategies (including the particle swarm test generator (PSTG) \cite{Ref8}, DPSO \cite{Ref9}, APSO \cite{Ref10}, and CS \cite{Ref11}) for the t-way test minimization problem.

\end{itemize}

\section{Preliminaries\label{Preliminaries}}

\subsection{Covering Array Notation}

The generation (and minimization) of combinatorial test suites from both practical and theoretical perspectives is currently an active research area. Theoretically, the combinatorial test suite depends on a well-known mathematical object called the covering array (CA). Originally, the CA gained more attention as a practical alternative to the oldest mathematical object, the orthogonal array (OA), which had been used for statistical experiments \cite{Ref12,Ref13}.

An $OA_\lambda (N; t, k, v)$ is an $N \times k$ array, where for every $N \times t$ sub-array, each $t-tuple$ occurs exactly $\lambda$ times, where $\lambda = N/{v^t}$; $t$ is the combination strength; $k$ is the number of input functions ($k \geqslant t$); and $v$ is the number of levels associated with each input parameter of the software-under-test (SUT) \cite{Ref14}. Practically, it is very hard to translate these firm rules except in small systems with few input parameters and values. Therefore, there is no significant benefit for medium- and large-scale SUT because it is very hard to generate $OA$s. In addition, based on the rules mentioned above, it is not possible to represent the $OA$ when each input parameter has different levels.

To address the limitation of the $OA$, the $CA$ was introduced. A $CA_\lambda (N; t, k, v)$ is an $N \times k$ array over ($0, \cdots, v-1$) such that every $t-tuple$ is $\lambda$-covered and every $N \times v$ sub-array contains all ordered subsets of size $t$ of $v$ values at least $\lambda$ times, where the set of columns is $B=\lbrace b_0, \cdots, b_{v-1} \rbrace \supseteq {0, \cdots, k-1}$ \cite{Ref15,Ref16}. In this case, each $tuple$ appears at least once in the $CA$. In summary, any covering array $CA (N; t, k, v)$ can also be expressed as $CA (N; t, v^k)$.

Variations in the number of component can be handled by a mixed covering array ($MCA$) \cite{Ref17}. An $MCA (N; t, k, (v_1, v_2, \cdots, v_k))$ is an $N \times k$ array on $v$ values, where the rows of each $N \times t$ sub-array cover all $t$ interactions of values from the $t$ columns that occur at least once. For more flexibility in the notation, the array can be represented by $MCA (N; t, v_1^{k_1} v_2^{k_2} \cdots v_k^k)$.

\subsection{Motivating Example}

\begin{figure}

\begin{center}
\includegraphics[width=0.7\linewidth]{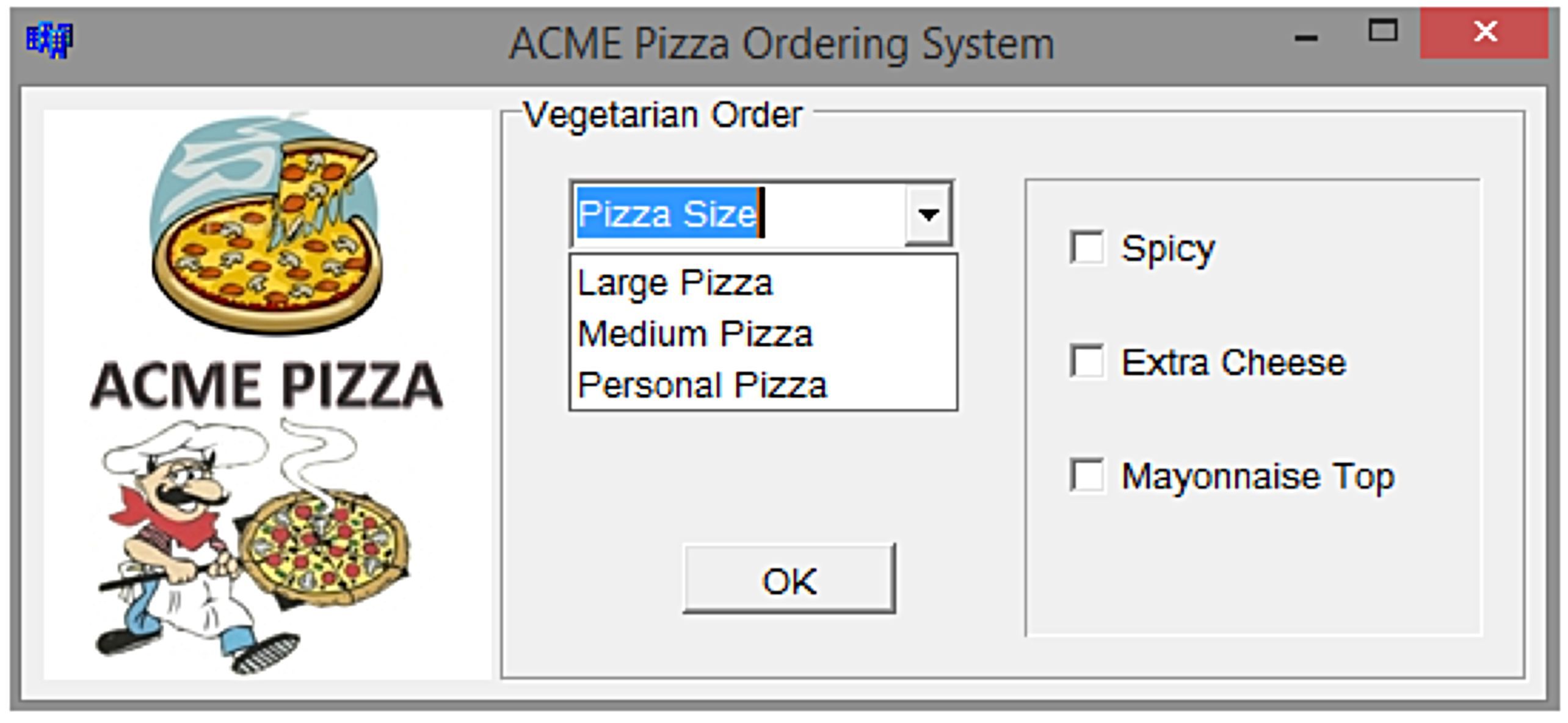}
\end{center}
\caption{Acme Vegetarian Pizza Order System}
\label{Figure1-Pizza}
\end{figure}

\begin{figure}

\begin{center}
\includegraphics[width=0.9\linewidth]{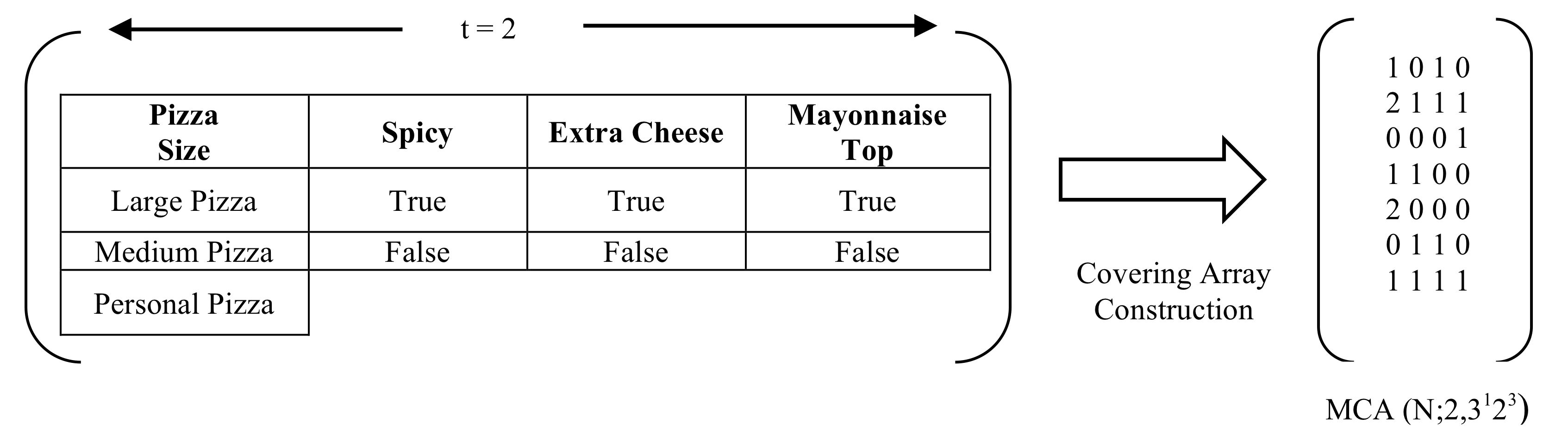}
\end{center}
\caption{Mixed Covering Array Construction $MCA (N;2,3^1 2^3)$ for Acme Vegetarian Pizza Order System}
\label{Figure2-Pizza_Table}
\end{figure}

To illustrate the use of the CA for $t-way$ testing, consider the hypothetical example of an Acme Vegetarian Pizza Ordering System. Referring to Fig \ref{Figure1-Pizza}, the system offers four selections of parameters: Pizza Size, Spicy, Extra Cheese, and Mayonnaise Topping. One parameter takes three possible values (Pizza Size = $\lbrace$ Large Pizza, Medium Pizza, and Personal Pizza$\rbrace$), while the rest of the parameters take two possible  values (Spicy = $\lbrace$ True, False$\rbrace$, Extra Cheese = $\lbrace$ True, False$\rbrace$, and Mayonnaise Topping = $\lbrace$ True, False$\rbrace$). Ideally, an all-exhaustive test combination requires $3 \times 2 \times 2 \times 2 = 24$ combinations. In a real-life testing scenario, the number of parameters and values can be enormous, resulting in a potentially large number tests. Given tight production deadlines and limited resources, test engineers can resort to a t-wise sampling strategy to minimize the test data for systematic testing. In the context of the Acme Vegetarian Pizza Ordering System highlighted earlier, the mixed-strength CA representation for $MCA (N;2, 3^1 2^3)$ can be seen in Fig \ref{Figure2-Pizza_Table} with 7 test cases (a reduction of 70.83\% from the 24 exhaustive possibilities). Table \ref{Table1:Mapping-of-MCA-Table} highlights the corresponding test cases mapped from the given mixed-strength covering arrays. Ideally, the selection of the previously mentioned (mixed) CA representation depends on the product requirements and the creativity of the test engineers based on the given testing.

\begin{table}

\caption{\label{Table1:Mapping-of-MCA-Table}Mapping of Mixed Covering Arrays
to Test Cases}

\begin{centering}
\begin{tabular}{|c|c|c|c|c|}
\hline 
\multicolumn{5}{|c|}{$MCA(N;2,3^{1}2^{3})$}\tabularnewline
\hline 
\hline 
Test ID & Pizza Size & Spicy & Extra Cheese  & Mayonnaise Top\tabularnewline
\hline 
1 & Medium Pizza & False & True  & False\tabularnewline
\hline 
2 & Large Pizza  & True  & True  & True\tabularnewline
\hline 
3 & Personal Pizza  & False & False & True\tabularnewline
\hline 
4 & Medium Pizza  & True & False & False\tabularnewline
\hline 
5 & Large Pizza  & False & False & False\tabularnewline
\hline 
6 & Personal Pizza  & True & True & False\tabularnewline
\hline 
7 & Medium Pizza  & True & True & True\tabularnewline
\hline 
\end{tabular}
\par\end{centering}
\end{table}

Mathematically, the $t-way$ test generation problem can be expressed by: 

\begin{equation}\
f(Z)=\left|\left\{ {I\:in\:VIL:Z\:covers\:I}\right\} \right| 
\label{Eq1}
\end{equation}

\begin{center}
$Subject\:to\:Z=Z_{1},Z_{2},\ldots,Z_{i}\:in\text{ }P_{1},P_{2},\ldots,P{}_{i}\:;\:i=1,2,\ldots,N$
\end{center}

where $f(Z)$ is an objective function (or the fitness function); $Z$ is the test case candidate, which is the set of decision variables $Z_i$; $VIL$ is the set of non-covered interaction $tuples (I)$; the vertical bars $| • \: |$ represent the cardinality of the set and the objective value is the number of non-covered interaction tuples covered by $Z$; $P_i$ is the set of possible values for each decision variable, with $P_i =$ discrete decision variables $(Z_i(1)< Z_i(2)<\cdots<Z_i(K))$; $N$ is the number of decision variables (here the parameters); and $K$ is the number of possible values for the discrete variables.

\section{Related Work}

As part of the general interest in search-based software engineering (SBSE) approaches \cite{Ref18}, much research attention has been given to the application of meta-heuristic search techniques to address the combinatorial test generation problem. Meta-heuristic techniques have had a big impact on the construction of $t-way$ and variable-strength test suites, especially in terms of the optimality of the test suite \cite{Ref19,Ref20,Ref21,Ref22,Ref23} .

Meta-heuristic based strategies often start with a population of random solutions. Then, one or more search operators are iteratively applied to the population to improve the overall fitness (greedily covering the interaction combinations). While there are many variations, the main difference between meta-heuristic strategies is based on each search operator and how exploration and exploitation are manipulated.

Cohen et al. \cite{Ref16,Ref24} developed a simulated annealing-based strategy for supporting the construction of a uniform and variable-strength $t-way$ test suite. A large random search space is generated in the implementation. When the algorithm iterates, the strategy chooses better test cases to construct the final test suite using the binary search process and a transformation equation. The search space is transformed from one state to another according to a probability equation. The results of the study are mainly concerned with the interaction strengths of two and three \cite{Ref24}.

Chen et al. \cite{Ref25} implemented a $t-way$ strategy based on ant colony optimization (ACO). The strategy simulates the behaviour of natural ant colonies in finding paths from the colony to the location of food. Each ant generates one candidate solution and walks through all paths in this solution by probabilistically choosing individual values. When the ant reaches the end of the last path, it returns and updates the initial candidate solution accordingly. This process continues until the iteration is complete. The final test case is chosen according to the maximum coverage of the t-interaction. Unlike the SA, the final test suite is further optimized by a merging algorithm that tries to merge the test cases.

Shiba et al. \cite{Ref26} adopted a genetic algorithm (GA) based on natural selection. Initially, the GA begins with randomly created test cases called chromosomes. These chromosomes undergo crossover and mutation until the termination criterion is met. In each cycle, the best chromosomes are probabilistically selected and added to the final test suite.

Alsewari et al. \cite{Ref19} developed a $t-way$ strategy based on the harmony search algorithm (HSS). The HSS mimics the behaviour of musicians trying to compose good music either by improvising on the best tune they remember or by random sampling. In doing so, the HSS iteratively exploits the harmonic memory to store the best solution found through a number of defined probabilistic improvisations within its local and global search processes. In each improvisation, one test case is selected for the final test suite until all the required interactions are covered. The notable feature of the HSS is that it supports constraints using the forbidden tuple approach.

Ahmed et al. \cite{Ref11} adopted the cuckoo search algorithm (CS), which mimics the unique lifestyle and aggressive reproduction strategy of the cuckoo. First, the CS generates random initial eggs in host nests. Each egg in a nest represents a vector solution that represents a test case. In each generation, two operations are performed. Initially, a new nest is generated (typically through a Lévy flight) and compared with the existing nests. The new nest replaces the current nest if it has a better objective function. Then, the CS adopts probabilistic elitism to maintain the elite solutions for the next generation.

Particle swarm optimization (PSO) \cite{Ref2} is perhaps the most popular implementation of $t-way$ test suite generation. The PSO-based $t-way$ strategy searches by mimicking the swarm behaviour of flocking birds. In PSO, global and local searches are guided by the inertia weight and social/cognitive parameters. Initially, a random swarm is created. Then, the PSO algorithm iteratively selects a candidate solution within the swarm to be added to the final test suite until all interaction tuples are covered (based on velocity and displacement transformation). Ahmed et al. developed early PSO-based strategies called the PSTG \cite{Ref8,Ref29} and APSO \cite{Ref10}. APSO is an improvement on the PSTG integrated with adaptive tuning based on the Mamdani fuzzy inference system \cite{Ref30,Ref31}. Wu et al. implemented discrete PSO \cite{Ref8} by substantially modifying the displacement and velocity transformation used in PSO. The benchmark results of DPSO \cite{Ref9} demonstrate its superior performance when compared with both the PSTG and APSO.

Despite the significant number of proposed algorithms in this field, the adoption of new meta-heuristic algorithms is most welcome. The no free lunch (NFL) theorem \cite{Ref32} suggests that no single meta-heuristic algorithm can outperform others even when there is a slight change in the problem of ($t-way$) configurations. Therefore, the NFL theorem allows researchers to propose new algorithms or modify current ones to enhance the current solution. In fact, the results could also be applied in other fields.

Hybrid integration with machine learning appears to be a viable approach to improving the state-of-the-art meta-heuristic algorithms. Machine learning relates to the study of the fundamental laws that govern the computer learning process concerning the construction of systems that can automatically learned from experience. Machine learning techniques can be classified into three types: supervised, unsupervised, and reinforcement \cite{Ref33}. Supervised learning involves learning a direct functional input-output mapping based on some set of training data and being able to predict new data. Unlike supervised learning, unsupervised learning does not require explicit training data. Specifically, unsupervised learning involves learning by drawing inferences (e.g., clustering) from an input dataset. Reinforcement learning allows mappings between states and actions to maximize reward signals using experimental discovery. This type of learning differs from supervised learning in that it relies on a punishment and reward mechanism and never corrects input-output pairs (even when dealing with suboptimal responses).

Combinatorial test suite minimization is one of the crucial elements of an efficient test design \cite{Ref34,Ref35}. This area is worth further exploration, especially when we can take advantage of machine learning’s benefits. To be specific, our approach focuses on the hybridization of a meta-heuristic algorithm with reinforcement learning based on the Q-learning algorithm \cite{Ref6}. The Q-learning algorithm is attractive due to its successful adoption in many prior works. Ant-Q \cite{Ref36} is the first attempt by researchers to integrate a meta-heuristic algorithm (ACO) with Q-learning. Although its integration with Q-learning is useful (e.g., it has been successfully adopted for the 2-dimensional cutting stock problem \cite{Ref37} and the nuclear reload problem \cite{Ref38}), the approach appears too specific to ACO because pheromones and evaporation are modelled as part of the Q-learning updates (as rewards and punishments). In a more recent study, RLPSO \cite{Ref39}, a PSO algorithm integrated with Q-learning, was successfully developed (and adopted in a selected case study of a gear and pressure vessel design problem and standard benchmark functions). While it has merit, the approach is computationally intensive and complex because each particle in the swarm must carry its own Q-metrics. Therefore, the RLPSO approach is not sufficiently scalable for large-scale combinatorial problems requiring large population sizes. In the current study, the swarm size is limited to 3.

By building on and complementing the work mentioned above, our work explores the hybridization of the Q-learning algorithm with a recently developed meta-heuristic algorithm called the SCA \cite{Ref5}. Unlike most meta-heuristic algorithms that mimic certain physical or natural phenomena, the equation transformation used in the SCA is solely based on the sine and cosine operations. Therefore, the learning curve of the SCA is low. Although its exploitation is commendable, the exploration of the SCA is strictly bounded due to the (adaptive) shrinking magnitude of the sine and cosine functions’ multipliers during the search process. To address the issues mentioned above, we propose a new algorithm, the QLSCA. Moreover, we augment the QLSCA with two further operations (Lévy flight motion and crossover) to counterbalance its exploration and exploitation. Then, we use the Q-learning algorithm (which is based on the penalty and reward mechanism) to dynamically identify on the best operation (sine, cosine, Lévy flight motion, or crossover) during runtime.

\section{Overview of the SCA}

The SCA is a population-based meta-heuristic algorithm \cite{Ref5}. As the name suggests, the SCA exploits the sine and cosine functions to update its population’s positions. Each position is treated as a vector. To be specific, the positions are updated is based on:

\begin{equation}
X_{i}^{(t+1)}=X_{i}^{t}+r_{1}\times sin(r_{2})\times \left|r_{3}P_{i}^{t}-X_{i}^{t}\right|,\;r_{4}<0.5
\end{equation}

\begin{equation}
X_{i}^{(t+1)}=X_{i}^{t}+r_{1}\times cos(r_{2})\times \left|r_{3}P_{i}^{t}-X_{i}^{t}\right|,\;r_{4}\geq0.5
\end{equation}

where $X_i^t$ is the position of the current solution in the $i^{th}$ dimension and the $t^{th}$ iteration; $r_1$, $r_2$, $r_3$, and $r_4$ are random numbers in [0,1]; $P_i$ is the position of the best destination point in the $i^{th}$ dimension, and $| • \: |$ indicates the absolute value.

\begin{figure}

\begin{center}
\includegraphics[width=0.9\linewidth]{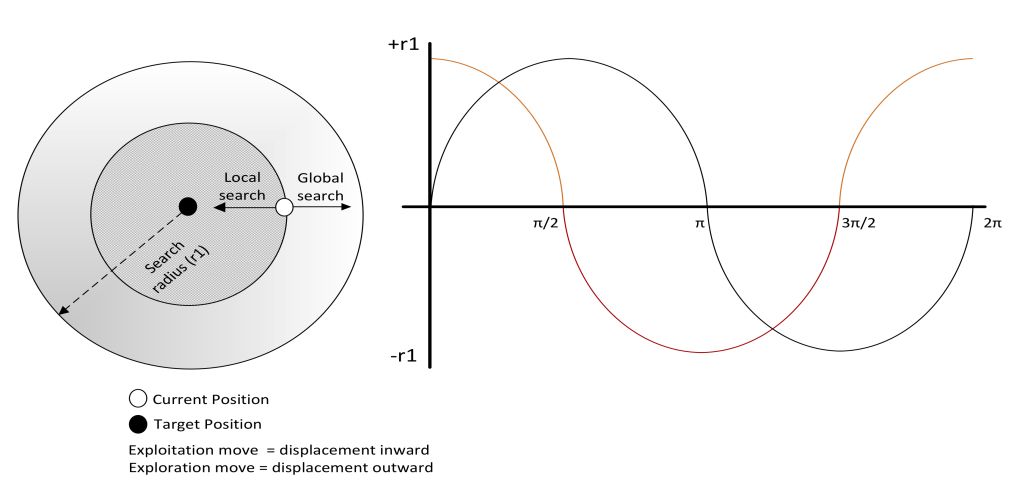}
\end{center}
\caption{Effects of Sine and Cosine on Search Radius}
\label{Figure3-Effect-of-Sin-cos-fig}
\end{figure}

Due to its importance to the exploration and exploitation of the SCA, the four main parameters $r_1$, $r_2$, $r_3$, and $r_4$ require further elaboration. The parameter $r_1$ dictates the radius of the search circle (displacement size). It is also possible to adaptively and dynamically vary $r_1$ during the iteration process using:

\begin{equation}
r_{i}=M(1-\frac{t}{T})
\end{equation}

where $t$ is the current iteration; $T$ is the maximum number of iterations; and $M$ is a constant. Due to the cyclic nature of sine and cosine, the parameter $r_2$ defines whether the motion is inward (the direction of exploitation when sine and cosine are negative) or outward (the direction of exploration when sine and cosine are positive), as can be seen in Fig \ref{Figure3-Effect-of-Sin-cos-fig}. The parameter $r_3$ brings in the random weight from the best position to affect the overall displacement from the current position. Finally, the parameter $r_4$ equally switches between the sine and cosine components.

To summarize, the general pseudo code for the SCA algorithm is given in Algorithm \ref{SCA-Algo}.

\begin{algorithm}   
\linespread{1.1}\selectfont

 \KwIn{the population $X = \lbrace X_1, X_2,\cdots ,X_D \rbrace$, the constant magnitude $M$}
 \KwOut{$X_{best}$ and the updated population $X^{\prime}= \lbrace X_1^{\prime}, X_2^{\prime},\cdots , X_N^{\prime} \rbrace$ }

Initialize random population $X$

\While {(stopping criteria not met)}{

Set initial $r_1$  using Eq.3

\For {iteration = 1 till  max iteration}{

\For{population count  =1 to population size}{

Evaluate each population of $X$ by the objective function

Update the best solution obtained so far,  $P_i^t= X_{best}$

Randomly generate the value of  $r_2$, $r_3$, $r_4$ between [0,1]\

Update the position of $X$ using Eq.2 or Eq.3

Update $r_1$ adaptively using Eq.4

}

}

}

 Return the updated population, $X$ and the best result ($X_{best}$)

 \caption{Pseudo Code for SCA Algorithm \cite{Ref5}\label{SCA-Algo}}
\end{algorithm}

\section{The Proposed Strategy }

The proposed QLSCA-based strategy integrates Q-learning with the sine and cosine operations, Lévy flight motion and crossover. Lévy flight and crossover were selected for two reasons. Firstly, the Lévy flight operator is a well-known global search operator \cite{Ref40}. Activating the Lévy flight operator can potentially propel the search process from a local optimum. Secondly, the crossover can be considered both global and local searching \cite{Ref1}. For instance, 1- or 2-point crossover can be regarded as local searching. However, crossover at more than 2 points is essentially global searching. Such flexible behaviour balances the intensification and diversification of the QLSCA.

Having justified the adoption of Lévy flight motion and crossover, the detailed explanation of the proposed QLSCA is as follows: 

\subsection{Q-Learning Algorithm}

The Q-learning algorithm \cite{Ref6} learns the optimal selection policy by interacting with the environment. The algorithm works by estimating the best state-action pair through the manipulation of a $Q(s, a)$ table. To be specific, a $Q(s, a)$ table uses a state-action pair to index a $Q$ value (as a cumulative reward). The $Q(s, a)$ table is dynamically updated based on the reward and punishment ($r$) of a particular state-action pair.

\begin{equation}
Q_{(t+1)}(s_{t},a_{t})=Q_{t}(s_{t},a_{t})+\alpha_{t}(r_{t}+\gamma max(Q_{t}(s_{(t+1)},a_{(t+1)}))-Q_{t}(s_{t},a_{t}))
\end{equation}

The optimal setting for $\alpha _t, \gamma$, and $r_t$ within the Q-learning algorithm requires further clarification. When $\alpha _t$ is close to 1, higher priority is given to the newly gained information for the Q-table updates. However, a small value of $\alpha _t$ gives higher priority to existing information. To facilitate exploration of the search space (to maximize learning from the environment), $\alpha _t$ can be set to a high value during early iterations and adaptively reduced in later iterations (to exploit the current best Q-value). This process is as follows:

\begin{equation}
\alpha_{t}=1-0.9\times\frac{t}{(Max\:Iteration)}
\end{equation}

The parameter $\gamma$ functions as a scaling factor for rewarding or punishing the Q-value based on the current action. When $\gamma$ is close to 0, the Q-value is based solely on the current reward or punishment. When $\gamma$ is close to 1, the Q-value is based on the current and previous reward and/or punishment. The literature suggests setting $\gamma = 0.8$ \cite{Ref39}.

The parameter $r_t$ serves as the actual reward or punishment. In our current study, the value of $r_t$ is set based on:

\begin{equation}
\begin{cases}
 & \begin{array}{c}
r_{t}=1,\:if\:the\:current\:action\:improves\:fitness\\
r_{t}=-1,otherwise
\end{array}\Biggr\}\end{cases}
\end{equation}

Summing up, the pseudo code of the Q-learning algorithm is illustrated in Algorithm \ref{Q-learning-Algo}.

\begin{algorithm}   
\linespread{1.1}\selectfont

 \KwIn{$S  = [s_1,s_2,\cdots ,s_n]$, $A = [a_1,a_2, \cdots ,a_n]$, $Q(s,a)$}
 \KwOut{Updated $Q(s,a)$ table }

Let $s_t$ be the state at a particular instance $t$
 
Let $a_t$ be the action at a particular instance $t$

 \For{each state $S  = [s_1,s_2,\cdots ,s_n]$ and action $A = [a_1,a_2, \cdots ,a_n]$}{
 
 Set $Qt(st,at)=0$
 
 }
 
 Randomly select an initial state, $s_t$
 
 \While{stopping criteria not met}{
 
 From the current state $s_t$, select the best action $a_t$ from the Q-table
 
 Execute action $a_t$  and get immediate reward/punishment $r_t$  using Eq.7
 
 Get the maximum Q value for the next state $s_{t+1}$
 
 Update $\alpha_t$  using Eq.6
 
 Update Q-table entry using  Eq.5
 
 Update the current state,  $s_t = s_{t+1}$

 }

 Return the updated $Q(s,a)$ table

 \caption{Pseudo Code for the Q-Learning Algorithm \label{Q-learning-Algo}}
\end{algorithm}

\subsection{Lévy Flight Motion}

\begin{figure}

\begin{center}
\includegraphics[width=0.9\linewidth]{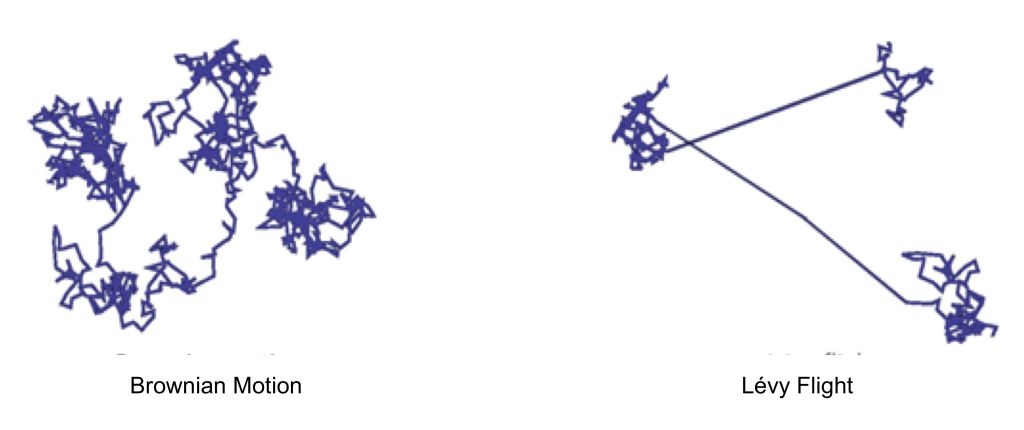}
\end{center}
\caption{Brownian Motion versus Lévy Flight Motion adopted from \cite{Ref41}}
\label{Figure4-Levy-Flight}
\end{figure}

To complement the sine and cosine operations within the SCA and ensure that the developed QLSCA can jump out of local minima, we propose incorporating Lévy flight. Yang popularizes Lévy flight motion in his implementation of the cuckoo search algorithm \cite{Ref7}. Essentially, a Lévy flight motion is a random walk (global search operation) that takes a sequence of jumps that are selected from a heavy-tailed probability function. Ideally, the jumps taken in a Lévy flight are unpredictable and consist of a mixture of extremely high and low displacements directed inward (negative) or outward (positive). As an illustration, Fig \ref{Figure4-Levy-Flight} compares a Lévy flight to a typical Brownian (random) walk.

Mathematically, the step length of a Lévy flight motion can be defined as follows \cite{Ref7}:

\begin{equation}
L\acute{e}vy\:Flight\,Step=\frac{u}{[v]^{(1/\beta)}}
\end{equation}

where $u$ and $v$ are approximated from a normal Gaussian distribution in which:

\begin{equation}
u\text{\ensuremath{\approx}}N(0,\sigma_{u}{}^{2})\text{\ensuremath{\centerdot}}\sigma_{u\;\;\quad}v\text{\ensuremath{\approx}}N(0,\sigma{}_{v}^{2})\text{\ensuremath{\centerdot}}\sigma_{v}
\end{equation}

For  $v$  value estimation, we use $\sigma_v=1$. For $u$ value estimation, we evaluate the Gamma function($ \Gamma $)  \cite{Ref42} with the value of $\beta=1.5$ \cite{Ref40}, and  obtain $\sigma_u$  using:

\begin{equation}
\sigma_{u}=\Biggl|\frac{\Gamma(1+\beta)\times sin(\frac{\pi\beta}{2})}{\Gamma(\frac{(1+\beta)}{2})\times\beta\times2^{\frac{(\beta-1)}{2}}}\Biggr|^{\frac{1}{\beta}}
\end{equation}

A lévy flight motion displacement update (with exclusive OR operation $\oplus$) is then defined as:

\begin{equation}
X_{i}^{(t+1)}=X_{i}^{t}\text{\ensuremath{\oplus}}L\acute{e}vy\;Flight\;Step
\end{equation}

\subsection{Crossover Operation}

\begin{figure}

\begin{center}
\includegraphics[width=0.9\linewidth]{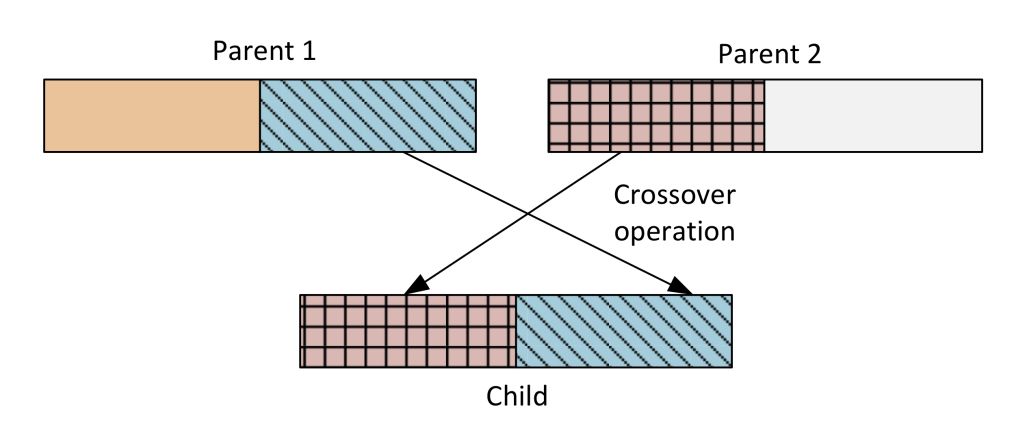}
\end{center}
\caption{Crossover Operation}
\label{Figure5-Cross-over-fig}
\end{figure}

The crossover operation is derived from GAs. Ideally, crossover is a local search operation whereby two distinct populations $X_i$ and $X_j$ exchange their partial values based on some random length $\beta$. Visually, crossover is represented in Fig \ref{Figure5-Cross-over-fig}.

The displacement due to crossover can be updated in the following three steps:

\begin{equation}
\begin{Bmatrix}Set\:crossover\:length\:\beta=random(0,lengthX_{i}^{t})\\
Generate\:random\;X{}_{j}^{t}\;such\;that\;i\neq j\\
Exchange\:from\:position\:0\;to\:position\;\beta\;from\;X{}_{j}^{t}\;to\;X{}_{i}^{t}
\end{Bmatrix}
\end{equation}

\subsection{The QLSCA Algorithm}

Exploration and exploitation are the key components of reinforcement learning algorithms (such as Q-learning). Exploration is necessary to understand the long-term rewards and punishment to be used later during exploitation of the search space. Often, it is desirable to explore during the early iterations. During later iterations, it is then desirable to favour exploitation. To achieve such an effect, Mauzo et al. suggest adopting the Metropolis probability function criterion \cite{Ref43}, which is mainly used in simulated annealing. Alternatively, a more straightforward probability criterion with a similar effect (decreasing over time) can be defined as:

\begin{equation}
\begin{array}{c}
Random\:r[0,1]\;<\:\frac{1}{\sqrt{iteration}}\;go\:to\:exploration\:mode\\
Otherwise,\;go\:to\:exploitation\:mode
\end{array}\Biggr\}
\end{equation}

Our early experiments with the Metropolis probability function indicate no significant performance difference with the probability given in Eq. 13. Furthermore, the Metropolis probability function’s exploitation of the current and previous values can be problematic, since Q-learning is a Markov decision process that relies on the current and forward-looking Q-values. For these reasons, we do not favour Metropolis-like probability functions.

\begin{figure}

\begin{center}
\includegraphics[width=0.9\linewidth]{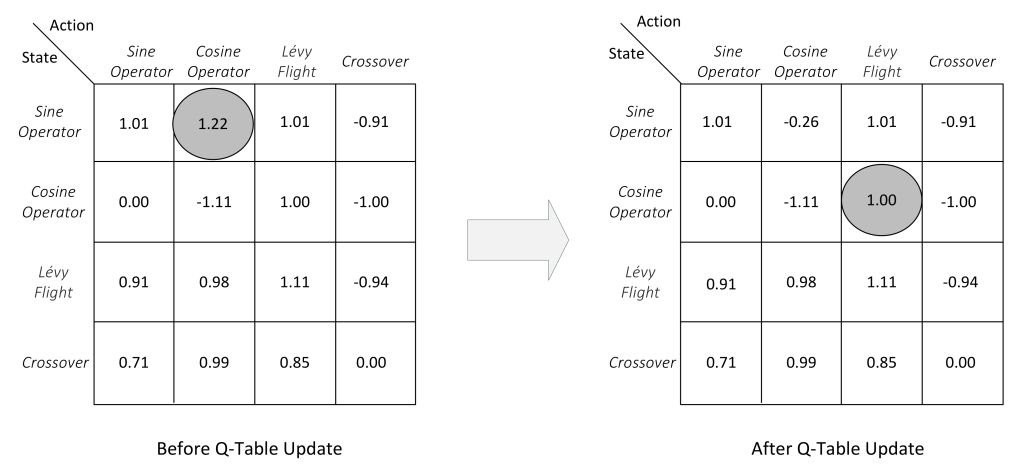}
\end{center}
\caption{Q-Table Update}
\label{Figure6-Q-Table-fig}
\end{figure}

To ensure that the learning is adequate (i.e., the roaming of the search space is sufficient), the QLSCA updates the Q-table for one complete episode cycle (in some random order) for each exploration opportunity. To support the use of 4 search operators within the QLSCA (sine, cosine, Lévy flight and crossover), the Q-table needs to be constructed as a 4×4 matrix in which the rows represent the state ($s_t$) and the columns represent the action ($a_t$) for each state. Fig \ref{Figure6-Q-Table-fig} depicts a snapshot of the Q-table for the QLSCA along with a numerical example. Assume that the current state-action pair is $s_t$ = $Sine \: Operator$ and has $a_t$ = $Cosine \: Operator$. The search process selects one of the four operators (sine, cosine, Lévy flight, and crossover) as the next action ($a_t$) based on the maximum reward defined in the state-action pair within the Q-table. This is unlike the original SCA algorithm in that the cosine or sine operator is selected based on the probability parameter, $r_4$.

Referring to Fig \ref{Figure6-Q-Table-fig}, we assume that the settings are as follows: the current value stored in the Q-table for the current state is $Q_{(t+1)} (s_t,a_t )= 1.22$ (grey circle in Fig \ref{Figure6-Q-Table-fig}); the reward is $r_t = -1.00$; the discount factor is $\gamma = 0.10$; and the current learning factor is $a_t$= 0.70. Then, the new value for $Q_{(t+1)} (s_t, a_t)$ in the Q-table is updated based on Eq. 4 as follows:

$Q_{(t+1)} (s_t,a_t )=1.22+0.70 \ast [-1.00+0.10 \ast Max(0.00,-1.11,1.00,-1.00)-1.22]=-0.26$

The state is then changed from sine to cosine. Similarly, the action $a_t= Cosine \: Operator$ is changed to Lévy flight. It should be noted that during both the exploration and exploitation of Q-value updates, the meta-heuristic QLSCA search process continues in the background. In other words, for each update, $X_{best}$ is kept and the population $X$ is updated accordingly.

Finally, based on the adoption of Lévy flight with sporadic long jumps, the positional update may sometimes encounter out-of-range values. Within the QLSCA, we establish a \textit{clamping rule} to apply lower and upper bounds to parameter values. In this way, when $X_j$ moves out of range, the boundary condition brings it back into the search space. There are at least three types of boundary condition: invisible walls, reflecting walls, and absorbing walls \cite{Ref44}. With invisible walls, when a current value goes beyond the boundary, the corresponding fitness value is not computed. With reflecting walls, when a value reaches the boundary, its value is reflected back to the search space (mirroring effects). With absorbing walls, when a current value moves out of range, the boundary condition brings it back into the search space by moving it to another endpoint. For example, if the position of a parameter limited to the range from 1 to 4, then, when the position exceeds 4, it is reset to 1. For our QLSCA implementation, we favour absorbing walls for our clamping rule.

To summarize, the complete QLSCA algorithm can be described in three main steps (Step A: Initialization, Step B: Exploration to Update the Q-table, and Step C: Exploitation to Update the Q-table), as shown in Algorithm \ref{Figure9-QLSCA Algorithm Figure 9}. As the name suggests, Step A involves initialization. Step B includes a complete update of the state-action pair update in 1 cycle in random order. Finally, Step C updates the currently selected state-action pair.

\begin{algorithm}
\scriptsize

 \KwIn{$S = \lbrace s_1, s_2, \cdots , s_n \rbrace$, $A = \lbrace a_1, a_2,\cdots , a_n \rbrace $, $Q(s, a)$, population $X = \lbrace X_1, X_2, \cdots , X_D\rbrace$, constant magnitude M,  Interaction strength ($t$), parameters $k =\lbrace k_1, k_2, \cdots , k_n \rbrace$,  values for each parameters $k, v =\lbrace v_1,v_2,\cdots ,v_n \rbrace$}
 
 \KwOut{Updated $Q(s, a)$ table, $X_{best}$, updated population $X^\prime = \lbrace X_1^\prime , X_2^\prime , \cdots ,  X_N^\prime \rbrace$, Final test suite $F_s$ }
 
 \tcc{Step A: Initialization}
 
Let $s_t$ be the state at a particular instance $t$
 
Let $a_t$ be the action at a particular instance $t$
 
 \For {each state $S = \lbrace s_1, s_2, \cdots , s_n \rbrace$ and action $A = \lbrace a_1, a_2,\cdots , a_n \rbrace$ }{
 
 Set $Q_t(s_t, a_t) = 0$ 
 
 }
 
Generate interaction tuple list based on the values of  $t, k, v$ (refer to Fig \ref{Figure7-HashTableGraph})

Randomly select an initial state, $s_t$

\While {interaction tuple list is not empty}{

\For {iteration = 1 till  max iteration}{

\For {population count  =1 till population size}{

Set initial $r_1$  using Eq.3

Choose Step B or Step C probabilistically based on Eq.12 

\tcc{Step B: Exploration for Q-table update}

\For {each state $S = \lbrace s_1, s_2, \cdots , s_n \rbrace$, and action $A = \lbrace a_1, a_2…, a_n \rbrace$ in random order \tcp*{loop for 1 episode}} 
{
From the current state $s_t$, select the best action $a_t$ from the Q-table \label{step12}

\If {action ($a_t$) == Sine operation }{
update $X_i^t$ using Eq.2}

\ElseIf{action ($a_t$) == Cosine operation}
{
update $X_i^t$ using Eq.3
}

\ElseIf{action ($a_t$) == Lévy flight motion}{
update $X_i^t$ using Eq.11
}

\ElseIf{action ($a_t$) == Crossover operation}{
update $X_i^t$ using Eq.12
}

Execute action $a_t$  and get immediate reward/punishment $r_t$  using Eq.7

Get the maximum Q value for the next state $s_{t+1}$

Update $a_t$  using Eq.6 

Update Q-table entry using  Eq.5

Update the current state,  $s_t = s_{t+1}$

Update $r_1$ adaptively using Eq.4

Update the best solution obtained so far,  $P_i^t= X_{best}$ \label{step24}

}

\tcc{Step C: Exploitation for Q-table update}

Redo Steps \ref{step12}-\ref{step24} \tcp* {loop for 1 complete episode unnecessary} 

}

}

Obtain the best result ($X_{best}$) from the updated population, $X$

}

 \caption{Pseudo Code for the QLSCA Algorithm \label{Figure9-QLSCA Algorithm Figure 9}}
\end{algorithm}

\section{QLSCA Strategy for $t-way$ Test Suite Generation}

Having described the QLSCA algorithm, the following section outlines its use in addressing the $t-way$ test suite generation problem. In general, the QLSCA is a composition of two main algorithms: (1) an algorithm for generating interaction tuples that generates combinations of parameters that are used in the test suite generator for optimization purposes, and (2) a QLSCA-based test suite generator algorithm. In the next sections, these two algorithms are detailed.

\subsection{Interaction Tuples Generation Algorithm}

The interaction tuples generation algorithm involves generating the parameter ($P$) combinations and the values ($v$) for each parameter combination. The parameter generation processes use binary digits: 0 indicates that the corresponding parameter is excluded and 1 indicates that it is included.

\begin{figure}
\begin{center}
\includegraphics[width=0.9\linewidth]{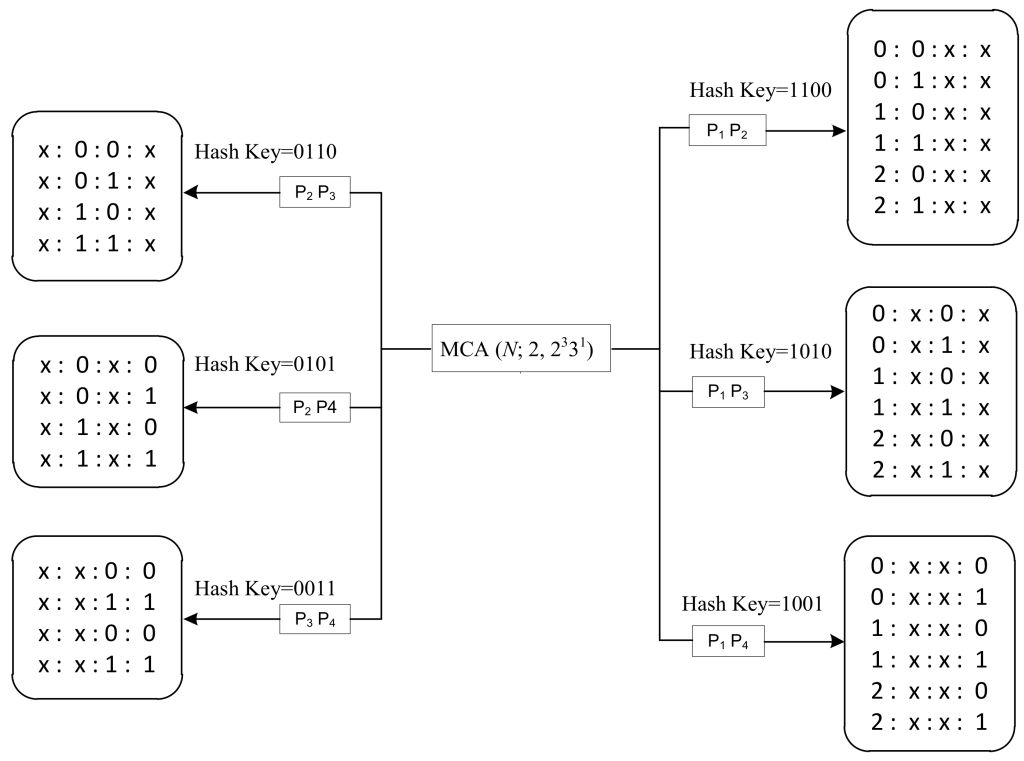}
\end{center}
\caption{The Hash Map and Interaction Tuples for $MCA (N; 2, 2^3 3^1)$}
\label{Figure7-HashTableGraph}
\end{figure}

Consider an example involving $MCA (N; 2, 2^3 3^1)$, as shown in Fig \ref{Figure7-HashTableGraph}. This configuration requires a $2-way$ interaction for a system of four parameters. First, the algorithm generates all possible binary numbers with up to four digits because there are 4 parameters. From these possibilities, the binary numbers that contain two “1”s are selected; these indicate that there is a pairwise interaction of parameters ($t = 2$). For example, the binary number 1100 refers to a $P_1P_2$ interaction. $P_1$ has two values (0 and 1), $P_2$ has two values (0 and 1), $P_3$ has two values (0 and 1), and $P_4$ has three values (0, 1, and 2). The $2-way$ parameter interaction has six possible combinations based on the parameter generation algorithm. For combination 1001, in which $P_1$ and $P_4$ are available, there are $2 \times 3$ possible interactions between $P_1$ and $P_4$. For each parameter in the combination (with two “1”s), the value of the corresponding parameter is included in the interaction elements. In this example, the excluded values are marked as “do not matter”. This process is repeated for the other five interactions, ($P_1, P_2$), ($P_1, P_3$), ($P_2, P_3$), ($P_2, P_4$), and ($P_3, P_4$).

To ensure efficient indexing for storage and retrieval, we opted to implement am interaction tuple hash table ($H_s$) that uses the binary representation of the interaction as the key. The complete algorithm for generating the interaction elements is highlighted in Algorithm \ref{Figure11-TupleGeneration-Algo-Figure11}.

\begin{algorithm}  
\linespread{1.1}\selectfont

 \KwIn{parameters’ values $p$, strength of coverage $t$}
 
 \KwOut{$H_s$ as the hash map of $t-way$ interaction tuples}
 
 Initialize $H_s = \lbrace \rbrace$
 
 Let m = max no of defined parameters
 
 Let $p = \lbrace p_0, p_1,\cdots , p_j\rbrace $, where $p$ represents the set of values defined for each parameter
 
 \For {index=0 to $2^{m – 1}$ }{
Let b = binary number

b = convert index to binary

\If {b[index] is equal  to 1}
{
Add the representitave interaction value for p[index]
} 
 
 \Else {Add don’t care value for p[index]}
 
 \If {the no of ‘1’s in b == t}{
 
 Set $hash_key$ = b
 
 Append the rest of p[index] with don’t care if necessary
 
 Put p into the hashmap, $H_s$, using the hash key

 }
 
 }
 
Return $H_s$

 \caption{Algorithm for Interaction Tuples Generation \label{Figure11-TupleGeneration-Algo-Figure11} }
\end{algorithm}

\subsection{Test Suite Generation Algorithm}

The principle underlying the QLSCA-based strategy is highlighted in Fig \ref{Figure6-Q-Table-fig}. Nevertheless, to apply the general QLSCA to the $t-way$ test generation problem, three adaptations must be made.

The first adaptation involves the input parameters. To cater to the $t-way$ problem, the QLSCA needs to process the parameters ($k$), the values ($v$) of each parameter, and the interaction strength ($t$). Based on these values, the interaction tuples can be generated.

The second adaptation is based on the way the population is represented within the QLSCA. The $t-way$ test generation problem is a discrete combinatorial problem. Therefore, the QLSCA initializes the population search space as a D-dimensional integer population $X_j = \lbrace X_{j,1}, X_{j,2}, X_{j,3}, \cdots, X_{j,D}]$ in which each dimension represents a parameter and contains integers between 0 and ($v_i$), which is the number of values the $i^{th}$ parameter takes.

Finally, the third adaptation is to the stopping criterion. When any particular interaction tuple has been covered (and the test case covering those tuples has been added to the final test suite $F_s$), then, the tuples are deleted from the interaction tuples list (refer to step \ref{Step30Figure12} in Algorithm \ref{Figure12-QLSCA-Strategy-Figure12}). Therefore, the QLSCA stops when the interaction tuple is empty (refer to step \ref{Step7figure12} in Algorithm \ref{Figure12-QLSCA-Strategy-Figure12}). The complete test suite generation algorithm based on the QLSA is summarized in Algorithm \ref{Figure12-QLSCA-Strategy-Figure12}.

\begin{algorithm}
\scriptsize
  
\linespread{1}\selectfont

 \KwIn{$S = \lbrace s_1, s_2, \cdots , s_n \rbrace$, $A = \lbrace a_1, a_2,\cdots , a_n \rbrace $, $Q(s, a)$, population $X = \lbrace X_1, X_2, \cdots , X_D\rbrace$, constant magnitude M,  \framebox{Interaction strength ($t$),} \framebox{parameters $k =\lbrace k_1, k_2, \cdots , k_n \rbrace$,  values for each parameters} \framebox{ $k, v =\lbrace  v_1,v_2,\cdots ,v_n \rbrace$}}
 
 \KwOut{Updated $Q(s, a)$ table, $X_{best}$, updated population $X^\prime = \lbrace X_1^\prime , X_2^\prime , \cdots ,  X_N^\prime \rbrace$, Final test suite $F_s$ }

\tcc{Step A: Initialization}

\For {each state $S = \lbrace s_1, s_2, \cdots , s_n \rbrace$ and action $A = \lbrace a_1, a_2,\cdots , a_n \rbrace $}{

Set $Q_t(s_t, a_t) = 0$ 
}

\framebox{Generate interaction tuple list based on the values of  $t, k, v$ (refer to Fig \ref{Figure7-HashTableGraph})}

Randomly select an initial state, $s_t$

\While {\framebox{interaction tuple list is not empty} \label{Step7figure12}}{

\For {iteration = 1 till  max iteration}{

\For {population count  =1 till population size}{

Set initial $r_1$  using Eq.3

Choose Step B or Step C probabilistically based on Eq.12

\tcc{Step B: Exploration for Q-table update}

\For {each state $S = \lbrace s_1, s_2, \cdots , s_n \rbrace$ and action $A = \lbrace a_1, a_2,\cdots , a_n \rbrace $ in random order \tcp*{ loop for 1 episode}}{

From the current state $s_t$, select the best action $a_t$ from the Q-table

\If{action ($a_t$) == Sine operation }{

update $X_i^t$ using Eq.1
}

\ElseIf {action ($a_t$) == Cosine operation}{
update $X_i^t$ using Eq.2
}

\ElseIf {action ($a_t$) == Lévy flight motion}{
update $X_i^t$ using Eq.10
}

\ElseIf {action ($a_t$) == Crossover operation}{
update $X_i^t$ using Eq.11
}

Execute action $a_t$  and get immediate reward/punishment $r_t$  using Eq.6

Get the maximum Q value for the next state $s_{t+1}$ 
	                 
Update $a_t$  using Eq.5

Update Q-table entry using  Eq. 4

Update the current state,  $s_t = s_{t+1}$
	               
Update $r_1$ adaptively using Eq.3
               
Update the best solution obtained so far,  $P_i^t= X_{best}$

}

\tcc{Step C: Exploitation for Q-table update}

Redo Steps \ref{step12}-\ref{step24} \tcp* {loop for 1 complete episode unnecessary}

}

Obtain the best result ($X_{best}$) from the updated population, $X$ 

\framebox{Add $X_{best}$ in the final test suite list, $F_s$ and delete the covered tuples in} \framebox{the interaction tuple list}\label{Step30Figure12}

}

}

 \caption{The QLSCA Strategy \label{Figure12-QLSCA-Strategy-Figure12} }
\end{algorithm}

\section{Experimental Study}

Our experiments focus on two related goals: (1) to characterize the performance of the QLSCA in comparison to that of the SCA, and (2) to benchmark the QLSCA and the SCA against other meta-heuristic approaches.

To achieve these goals, we have divided our experiments into three parts. In the first part, we run 3 selected CAs ($CA(N; 2, 3^{13})$, $CA(N; 2, 10^5)$, and $CA(N; 3, 4^6)$) and 3 selected MCAs ($MCA(N; 2, 5^1 3^8 2^2)$ , $MCA(N, 2, 6^1 5^1 4^6 3^8 2^3)$, and $MCA(N, 2, 7^1 6^1 5^1 4^6 3^8 2^3)$).

In the second part, we benchmark the sizes of the test suites generated by our SCA and QLSCA against those of existing meta-heuristics based on the benchmark $t-way$ experiments published in \cite{Ref9}. To be specific, the benchmark experiments involve the $CA (N; t, 3^k)$ with varying $t$ (from 2 to 4) and k (from 2 to 12), the $CA (N; t, v^7)$ and the $CA (N; t, v^{10})$ with varying $t$ (from 2 to 4) and $v$ (from 4 to 6).

In the third part, we analyse our results statistically. We intend to determine whether the performance of the QLSCA at minimizing the test suite size is a statistically significant improvement over compared to existing strategies.

We developed the SCA and the QLSCA using the Java programming language. For all experiments involving the SCA and the QLSCA, we set the population size = 40, max iterations = 100, and the constant M = 3 (refer to Eq. 3) for all the experiments. We execute the QLSCA and the SCA 30 times to ensure statistical significance. Our platform comprises a PC running Windows 10 with a 2.9 GHz Intel Core i5 CPU, 16 GB of 1867 MHz DDR3 RAM and a 512 MB flash HDD.

The best and mean times and sizes (whenever applicable) for each experiment are reported side-by-side. The best cell entries are marked with “$\star$”, while the best mean cell entries are in bold. Unavailable entries are denoted by NA.

To put our work into perspective, we highlight all the parameters for the strategies of interests (the PSTG \cite{Ref8}, DPSO \cite{Ref9}, APSO \cite{Ref10}, and the CS \cite{Ref11}) obtained from their respective publications (as depicted in Table \ref{Table2}).

\begin{table}

\caption{Algorithm Parameters for Strategies of Interestsn \label{Table2} }

\begin{tabular}{|l|l|l|}
\hline 
Strategies & Parameters & Values\tabularnewline
\hline 
\hline 
\multirow{4}{*}{PSTG} & Max Iteration & 100 \tabularnewline
\cline{2-3} 
 & Population Size & 80 \tabularnewline
\cline{2-3} 
 & Acceleration Coefficients ($c_{1}$ and $c_{2}$) & 1.375 \tabularnewline
\cline{2-3} 
 & Inertia Weight ($w$) & 0.3 \tabularnewline
\hline 
\multirow{7}{*}{DPSO} & Max Iteration & 250 \tabularnewline
\cline{2-3} 
 & Population Size & 80 \tabularnewline
\cline{2-3} 
 & Acceleration Coefficients ($c_{1}$ and $c_{2}$) & 1.3 \tabularnewline
\cline{2-3} 
 & Inertia Weight ($w$) & 0.5\tabularnewline
\cline{2-3} 
 & Probability Parameter 1 (pro1) & 0.5 \tabularnewline
\cline{2-3} 
 & Probability Parameter 2 (pro 2) & 0.3\tabularnewline
\cline{2-3} 
 & Probability Parameter 3 (pro 3) & 0.7 \tabularnewline
\hline 
\multirow{4}{*}{APSO} & Max Iteration & 100 \tabularnewline
\cline{2-3} 
 & Population Size & 80 \tabularnewline
\cline{2-3} 
 & Dynamic Acceleration Coefficients ($c_{1}$ and $c_{2}$) & $1\le c\le2$\tabularnewline
\cline{2-3} 
 & Dynamic Inertia Weight ($w$) & $1\le w\le2$\tabularnewline
\hline 
\multirow{3}{*}{CS} & Max Iteration & 100 \tabularnewline
\cline{2-3} 
 & Population Size & 100\tabularnewline
\cline{2-3} 
 & Probability $p_{a}$ &  0.25\tabularnewline
\hline 
\end{tabular}

\end{table}

\subsection{Characterizing the Performance of the SCA and the QLSCA}

This section highlights the experiments that compare the SCA and the QLSCA with respect to test suite size, execution time, consistency (i.e., the range of variation in the generated results) and convergence patterns of the SCA and the QLSCA. To objectively perform this comparison, both strategies adopt the same parameter settings and are implemented using the same data structure and implementation language.

\begin{figure}
\begin{center}
\includegraphics[width=0.99\linewidth]{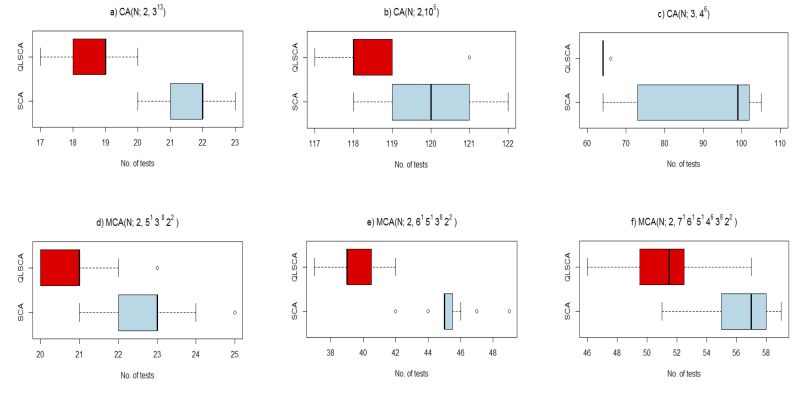}
\end{center}
\caption{Box Plots for Table \ref{Table3}}
\label{Figure8-BoxPlot}
\end{figure}

\begin{figure}
\begin{center}
\includegraphics[width=0.99\linewidth]{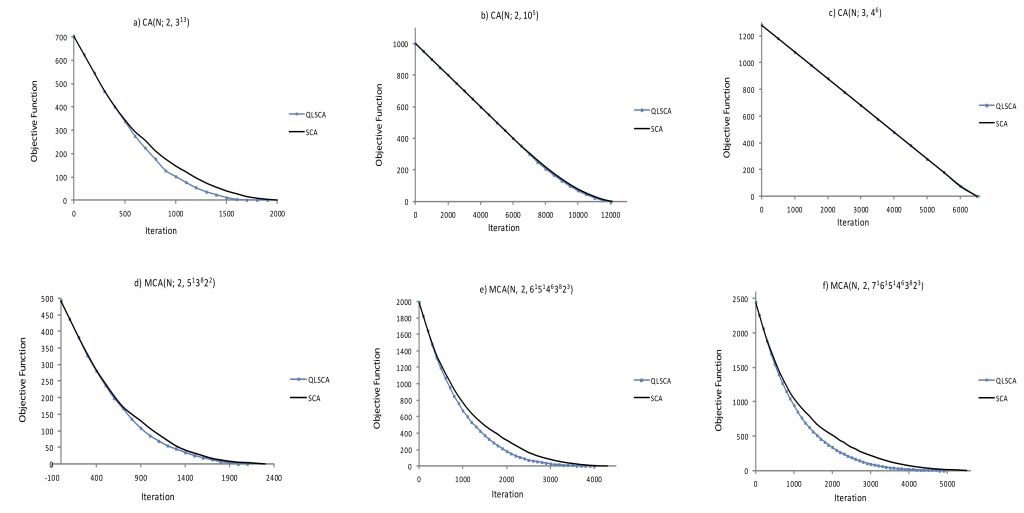}
\end{center}
\caption{Convergence Patterns for Table \ref{Table3}}
\label{Figure9-convergence-pattern}
\end{figure}

\begin{figure}
\begin{center}
\includegraphics[width=0.99\linewidth]{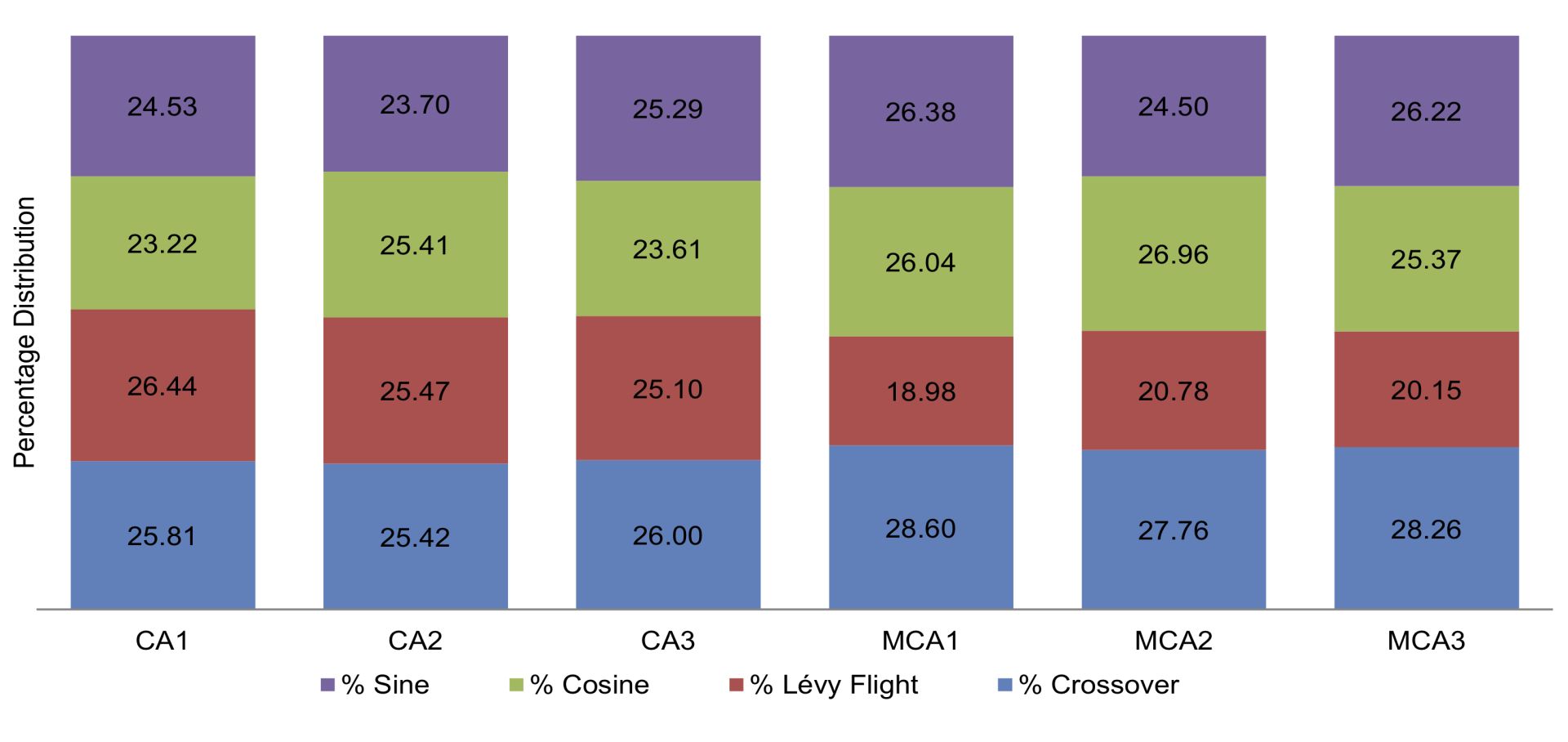}
\end{center}
\caption{Average Search Operator Percentage Distribution for  Table \ref{Table3}}
\label{Figure10-Bar-Figure}
\end{figure}

Table \ref{Table3} highlights our results for the test size and execution time. Fig \ref{Figure8-BoxPlot} depicts the box plot analysis. Fig \ref{Figure9-convergence-pattern} highlights the convergence patterns for the best 30 runs for each CA and MCA, while Fig \ref{Figure10-Bar-Figure} depicts the average percentage distribution for each search operator over all 30 runs.

\begin{table}

\caption{Time and Size Performances for SCA and QLSCA}
\scriptsize
\begin{tabular}{|l|c|c|c|c|c|c|c|c|}
\hline 
\multirow{3}{*}{CA and MCA} & \multicolumn{4}{c|}{SCA} & \multicolumn{4}{c|}{QLSCA}\tabularnewline
\cline{2-9} 
 & \multicolumn{2}{c|}{Size} & \multicolumn{2}{c|}{Time (sec)} & \multicolumn{2}{c|}{Size} & \multicolumn{2}{c|}{Time (sec)}\tabularnewline
\cline{2-9} 
 & Best & Mean & Best & Mean & Best & Mean & Best & Mean\tabularnewline
\hline 
$CA(N;2,3^{13})$ & 20  & 21.45 & 44.58{*}  & \textbf{51.04} & 17{*}  & \textbf{18.45} & 56.68  & 67.14\tabularnewline
\hline 
$CA(N;2,10^{5})$ & 118  & 120.10 & 19.55{*} & \textbf{20.55} & 117{*}  & \textbf{118.45} & 52.06  & 54.22\tabularnewline
\hline 
$CA(N;3,4^{6})$ & 64{*}  & 89.95 & 19.89{*}  & \textbf{29.27} & 64{*}  & \textbf{66.70} & 50.34  & 54.08\tabularnewline
\hline 
$MCA(N;2,5^{1}3^{8}2^{2})$ & 21  & 22.75 & 21.27{*}  & \textbf{24.01} & 20{*}  & \textbf{21.00} & 45.06  & 52.56\tabularnewline
\hline 
$MCA(N,2,6^{1}5^{1}4^{6}3^{8}2^{3})$ & 42  & 45.10 & 235.18{*} & \textbf{254.45} & 37{*} & \textbf{39.65} & 416.54  & 497.42\tabularnewline
\hline 
$MCA(N,2,7^{1}6^{1}5^{1}4^{6}3^{8}2^{3})$ & 51  & 56.35 & 299.80{*} & \textbf{358.60} & 46{*} & \textbf{51.25} & 634.02  & 940.24\tabularnewline
\hline 
\end{tabular}

\label{Table3}

\end{table}

\subsection{Benchmarking with other Meta-Heuristic based Strategies}

Unlike the experiments in the previous section, the benchmark experiments in this section (as adopted from \cite{Ref9}) also include comparisons of the QLSCA’s and SCA’s performances to those of all other strategies. However, the execution times have been omitted due to differences in the running environment and in the parameter settings (e.g., PSO relies on the population size, the inertia weight, and social and cognitive parameters, while the cuckoo search relies on the elitism probability, number of iterations, and population size) and implementation (e.g., the data structure and the implementation language). Tables \ref{Table3} to 10 highlight our complete results.

\begin{table}

\caption{Size Performance for $CA (N; 2, 3^k)$ where $k$ is varied from 3 to 12}
\scriptsize

\begin{tabular}{|c|c|c|c|c|c|c|c|c|c|c|c|c|}
\hline 
\multirow{2}{*}{K} & \multicolumn{2}{c|}{PSTG {[}8{]}} & \multicolumn{2}{c|}{DPSO {[}9{]}} & \multicolumn{2}{c|}{APSO {[}10{]}} & \multicolumn{2}{c|}{CS {[}11{]}} & \multicolumn{2}{c|}{SCA} & \multicolumn{2}{c|}{QLSCA}\tabularnewline
\cline{2-13} 
 & Best & Mean & Best & Mean & Best & Mean & Best & Mean & Best & Mean & Best & Mean\tabularnewline
\hline 
3 & 9{*} & 9.55 & NA  & NA & 9{*} & 9.21 & 9{*}  & 9.60  & 9{*}  & 9.70  & 9{*}  & 9.67\tabularnewline
\hline 
4 & 9{*}  & 10.15 & 9{*} & \textbf{9.00} & 9{*} & 9.95 & 9{*}  & 10.00 & 9{*} & 9.07 & 9{*} & \textbf{9.00}\tabularnewline
\hline 
5 & 12  & 13.81 & 11{*} & 11.53 & 11{*} & 12.23 & 11{*}  & 11.80 & 11{*} & 11.93 & 11{*} & \textbf{11.06}\tabularnewline
\hline 
6 & 13  & 15.11 & 14  & 14.50 & 12{*} & \textbf{13.78} & 13  & 14.20 & 13  & 14.10 & 14  & 14.27\tabularnewline
\hline 
7 & 15  & 16.94 & 15  & 15.17 & 15  & 16.62 & 14{*}  & 15.60 & 14{*} & 15.33 & 14{*}  & \textbf{15.10}\tabularnewline
\hline 
8 & 15{*}  & 17.57 & 15{*} & 16.00 & 15{*} & 16.92 & 15{*} & 15.80 & 5{*} & 16.33 & 15{*} & \textbf{15.77}\tabularnewline
\hline 
9 & 17  & 19.38 & 15{*}  & 16.43 & 16  & 18.31 & 16  & 17.20 & 16  & 17.78 & 5{*} & \textbf{16.27}\tabularnewline
\hline 
10 & 17  & 19.78 & 16  & 17.30 & 17  & 18.12 & 17  & 17.80 & 18  & 18.60 & 15{*} & \textbf{16.97}\tabularnewline
\hline 
11 & 17  & 20.16 & 17  & \textbf{17.70} & NA & NA & 18  & 18.60 & 18  & 19.80 & 16{*} & \textbf{17.70}\tabularnewline
\hline 
12 & 18  & 21.34 & 16{*}  & 17.93 & NA & NA & 18  & 18.80 & 19  & 20.33 & 16{*}  & \textbf{17.87}\tabularnewline
\hline 
\end{tabular}

\label{Table4}

\end{table}

\begin{table}

\caption{Size Performance for $CA (N; 3, 3^k)$ where $k$ is varied from 4 to 12}
\scriptsize

\begin{tabular}{|c|c|c|c|c|c|c|c|c|c|c|c|c|}
\hline 
\multirow{2}{*}{K} & \multicolumn{2}{c|}{PSTG {[}8{]}} & \multicolumn{2}{c|}{DPSO {[}9{]}} & \multicolumn{2}{c|}{APSO {[}10{]}} & \multicolumn{2}{c|}{CS {[}11{]}} & \multicolumn{2}{c|}{SCA} & \multicolumn{2}{c|}{QLSCA}\tabularnewline
\cline{2-13} 
 & Best & Mean & Best & Mean & Best & Mean & Best & Mean & Best & Mean & Best & Mean\tabularnewline
\hline 
4 & 27{*} & 29.30 & NA & NA & 27{*}  & 28.90 & 28  & \textbf{29.00} & 27  & 29.57 & 27{*}  & 29.70\tabularnewline
\hline 
5 & 39  & 41.37 & 41  & 43.17 & 41  & 42.20 & 38{*}  & \textbf{39.20} & 39  & 42.43 & 39  & 41.90\tabularnewline
\hline 
6 & 45  & 46.76 & 33{*}  & 38.30 & 45  & 46.51 & 43  & 44.20 & 33{*}  & 40.47 & 33{*}  & \textbf{37.57}\tabularnewline
\hline 
7 & 50  & 52.20 & 48{*}  & 50.43 & 48{*}  & 51.12 & 48{*} & 50.40 & 50  & 51.30 & 49  & \textbf{50.30}\tabularnewline
\hline 
8 & 54  & 56.76 & 52  & 53.83 & 50{*}  & 54.86 & 53  & 54.80 & 54  & 56.57 & 52  & \textbf{53.43}\tabularnewline
\hline 
9 & 58  & 60.30 & 56{*}  & 57.77 & 59  & 60.21 & 58  & 59.80 & 59  & 62.63 & 56{*} & \textbf{56.60}\tabularnewline
\hline 
10 & 62  & 63.95 & 59{*}  & 60.87 & 63  & 64.33 & 62  & 63.60 & 64  & 68.30 & 59{*}  & \textbf{60.63}\tabularnewline
\hline 
11 & 64  & 65.68 & 63  & 63.97 & NA & NA & 66  & 68.20 & 70  & 74.2 & 62{*}  & \textbf{63.37}\tabularnewline
\hline 
12 & 67  & 68.23 & 65{*}  & 66.83 & NA & NA & 70  & 71.80 & 78  & 80.73 & 65{*}  & \textbf{66.13}\tabularnewline
\hline 
\end{tabular}

\label{Table5}

\end{table}

\begin{table}

\caption{Size Performance for $CA (N; 4, 3^k)$ where $k$ is varied from 5 to 12}
\scriptsize
\begin{tabular}{|c|c|c|c|c|c|c|c|c|c|c|c|c|}
\hline 
\multirow{2}{*}{K} & \multicolumn{2}{c|}{PSTG {[}8{]}} & \multicolumn{2}{c|}{DPSO {[}9{]}} & \multicolumn{2}{c|}{APSO {[}10{]}} & \multicolumn{2}{c|}{CS {[}11{]}} & \multicolumn{2}{c|}{SCA} & \multicolumn{2}{c|}{QLSCA}\tabularnewline
\cline{2-13} 
 & Best & Mean & Best & Mean & Best & Mean & Best & Mean & Best & Mean & Best & Mean\tabularnewline
\hline 
5 & 96  & 97.83 & NA & NA & 94  & 96.33 & 94  & 95.80 & 81{*} & 87.50 & 81{*} & 84.63\tabularnewline
\hline 
6 & 133  & 135.31 & 131  & 134.37 & 129{*} & 133.98 & 132  & 134.20 & 130  & 133.80 & 129{*}  & 133.77\tabularnewline
\hline 
7 & 155  & 158.12 & 150{*} & 155.23 & 154  & 157.42 & 154  & 156.80 & 153  & 156.23 & 150{*} & 154.13\tabularnewline
\hline 
8 & 175  & 176.94 & 171{*} & 175.60 & 178  & 179.70 & 173  & 174.80 & 174  & 179.10 & 172  & 174.67\tabularnewline
\hline 
9 & 195  & 198.72 & 187  & 192.27 & 190  & 194.13 & 195  & 197.80 & 196  & 202.83 & 186{*} & 187.63\tabularnewline
\hline 
10 & 210  & 212.71 & 206  & 219.07 & 214  & 212.21 & 211  & 212.20 & 221  & 228.57 & 205{*} & 207.73\tabularnewline
\hline 
11 & 222  & 226.59 & 221  & 224.27 & NA & NA & 229  & 231.00 & 243  & 253.95 & 220{*} & 222.40\tabularnewline
\hline 
12 & 244  & 248.97 & 237  & 239.85 & NA & NA & 253  & 255.80 & 262  & 277.77 & 233{*} & 236.77\tabularnewline
\hline 
\end{tabular}

\label{Table6}

\end{table}

\begin{table}

\caption{Size Performance for $CA (N; 2, v^7)$ where $v$ is varied from 2 to 7}
\scriptsize
\begin{tabular}{|c|c|c|c|c|c|c|c|c|c|c|c|c|}
\hline 
\multirow{2}{*}{$v$} & \multicolumn{2}{c|}{PSTG {[}8{]}} & \multicolumn{2}{c|}{DPSO {[}9{]}} & \multicolumn{2}{c|}{APSO {[}10{]}} & \multicolumn{2}{c|}{CS {[}11{]}} & \multicolumn{2}{c|}{SCA} & \multicolumn{2}{c|}{QLSCA}\tabularnewline
\cline{2-13} 
 & Best & Mean & Best & Mean & Best & Mean & Best & Mean & Best & Mean & Best & Mean\tabularnewline
\hline 
2 & 6  & 6.82 & 7  & 7.00 & 6{*}  & \textbf{6.73} & 6{*} & 6.80 & 7  & 7.10 & 7  & 7.00\tabularnewline
\hline 
3 & 15  & 15.23 & 14{*}  & \textbf{15.00} & 15  & 15.56 & 15  & 16.20 & 15  & 15.54 & 15  & 15.10\tabularnewline
\hline 
4 & 26  & 27.22 & 24  & 25.33 & 25  & 26.36 & 25  & 26.40 & 25  & 26.73 & 23{*}  & \textbf{24.77}\tabularnewline
\hline 
5 & 37  & 38.14 & 34{*}  & 35.47 & 35  & 37.92 & 37  & 38.60 & 39  & 41.07 & 34{*}  & \textbf{35.37}\tabularnewline
\hline 
6 & NA & NA & 47{*}  & 49.23 & NA & NA & NA & NA & 54  & 57.30 & 48  & \textbf{48.90}\tabularnewline
\hline 
7 & NA & NA & 64{*}  & 66.37 & NA & NA & NA & NA & 73  & 75.70 & 64{*} & \textbf{65.47}\tabularnewline
\hline 
\end{tabular}

\label{Table7}

\end{table}

\begin{table}

\caption{Size Performance for $CA (N; 3, v^7)$ where $v$ is varied from 2 to 7}

\scriptsize
\begin{tabular}{|c|c|c|c|c|c|c|c|c|c|c|c|c|}
\hline 
\multirow{2}{*}{$v$} & \multicolumn{2}{c|}{PSTG {[}8{]}} & \multicolumn{2}{c|}{DPSO {[}9{]}} & \multicolumn{2}{c|}{APSO {[}10{]}} & \multicolumn{2}{c|}{CS {[}11{]}} & \multicolumn{2}{c|}{SCA} & \multicolumn{2}{c|}{QLSCA}\tabularnewline
\cline{2-13} 
 & Best & Mean & Best & Mean & Best & Mean & Best & Mean & Best & Mean & Best & Mean\tabularnewline
\hline 
2 & 13  & \textbf{13.61} & 15  & 15.06 & 15  & 15.80 & 12{*}  & 13.80 & 13  & 15.47 & 15  & 15.07\tabularnewline
\hline 
3 & 50  & 51.75 & 49  & 50.60 & 48{*}  & 51.12 & 49  & 51.60 & 48{*}  & 50.93 & 49  & \textbf{50.37}\tabularnewline
\hline 
4 & 116  & 118.13 & 112{*}  & 115.27 & 118  & 120.41 & 117  & 118.40 & 118  & 122.03 & 112{*}  & \textbf{115.23}\tabularnewline
\hline 
5 & 225  & 227.21 & 216  & 219.20 & 239  & 243.29 & 223  & 225.40 & 235  & 239.50 & 215{*}  & \textbf{218.00}\tabularnewline
\hline 
6 & NA & NA & 365  & 370.57 & NA & NA & NA & NA & 405  & 411.50 & 364{*}  & \textbf{369.53}\tabularnewline
\hline 
7 & NA & NA & 574  & \textbf{577.67} & NA & NA & NA & NA & 637  & 651.37 & 573{*} & 577.90\tabularnewline
\hline 
\end{tabular}

\label{Table8}

\end{table}

\begin{table}

\caption{Size Performance for $CA (N; 4, v^7)$ where $v$ is varied from 2 to 7}
\scriptsize
\begin{tabular}{|c|c|c|c|c|c|c|c|c|c|c|c|c|}
\hline 
\multirow{2}{*}{$v$} & \multicolumn{2}{c|}{PSTG {[}8{]}} & \multicolumn{2}{c|}{DPSO {[}9{]}} & \multicolumn{2}{c|}{APSO {[}10{]}} & \multicolumn{2}{c|}{CS {[}11{]}} & \multicolumn{2}{c|}{SCA} & \multicolumn{2}{c|}{QLSCA}\tabularnewline
\cline{2-13} 
 & Best & Mean & Best & Mean & Best & Mean & Best & Mean & Best & Mean & Best & Mean\tabularnewline
\hline 
2 & 29  & 31.49 & 34  & 34.00 & 30  & 31.34 & 27{*}  & \textbf{29.60} & 29  & 31.27 & 31  & 31.13\tabularnewline
\hline 
3 & 155  & 157.77 & 150  & 154.73 & 153  & 155.2 & 155  & 156.80 & 153  & 156.33 & 149{*}  & \textbf{154.70}\tabularnewline
\hline 
4 & 487  & 489.91 & 472{*} & \textbf{481.53} & 472  & 478.9 & 487  & 490.20 & 487  & 493.73 & 477  & 483.67\tabularnewline
\hline 
5 & 1176  & 1180.63 & 1148{*}  & \textbf{1155.63} & 1162  & 1169.94 & 1171  & 1175.20 & 1185  & 1203.00 & 1150  & 1159.23\tabularnewline
\hline 
6 & NA & NA & 2341{*}  & \textbf{2357.73} & NA & NA & NA & NA & 2465  & 2496.05 & 2347  & 2359.50\tabularnewline
\hline 
7 & NA & NA & 4290{*}  & \textbf{4309.60} & NA & NA & NA & NA & 4595  & 4618.40  & 4293  & 4315.00\tabularnewline
\hline 
\end{tabular}

\label{Table9}

\end{table}

\begin{table}

\caption{Size Performance for $CA (N; t, v^{10})$ where $t$ is varied from 2 to 4}
\scriptsize
\begin{tabular}{|c|c|c|c|c|c|c|c|c|c|c|c|}
\hline 
\multirow{2}{*}{$t$} & \multirow{2}{*}{$v$} & \multicolumn{2}{c|}{PSTG {[}8{]}} & \multicolumn{2}{c|}{DPSO {[}9{]}} & \multicolumn{2}{c|}{CS {[}11{]}} & \multicolumn{2}{c|}{SCA} & \multicolumn{2}{c|}{QLSCA}\tabularnewline
\cline{3-12} 
 &  & Best & Mean & Best & Mean & Best & Mean & Best & Mean & Best & Mean\tabularnewline
\hline 
\multirow{3}{*}{2} & 4 & NA  & NA & 28{*} & 29.20 & NA  & NA & 32  & 33.17 & 28{*}  & \textbf{28.63}\tabularnewline
\cline{2-12} 
 & 5 & 45  & 48.31 & 42  & 43.67 & 45  & 47.8 & 50  & 51.43 & 41{*}  & \textbf{43.13}\tabularnewline
\cline{2-12} 
 & 6 & NA  & NA & 58{*}  & \textbf{59.23} & NA  & NA & 71  & 73.13 & 58{*}  & 59.47\tabularnewline
\hline 
\multirow{3}{*}{3} & 4 & NA  & NA & 141  & 143.70 & NA  & NA & 166  & 171.77 & 140{*}  & \textbf{142.50}\tabularnewline
\cline{2-12} 
 & 5 & 287  & 298.00 & 273{*}  & 276.20 & 297  & 299.20 & 335  & 343.33 & 273{*}  & \textbf{274.60}\tabularnewline
\cline{2-12} 
 & 6 & NA  & NA & 467  & 470.50 & NA  & NA & 584  & 596.40 & 463{*}  & \textbf{468.83}\tabularnewline
\hline 
\multirow{3}{*}{4} & 4 & NA  & NA & 664  & 667.00 & NA  & NA & 743  & 779.25 & 657{*}  & \textbf{661.33}\tabularnewline
\cline{2-12} 
 & 5 & 1716  & 1726.72 & 1618  & 1620.80 & 1731  & 1740.20 & 1762  & 1788.25 & 1607{*}  & \textbf{1613.00}\tabularnewline
\cline{2-12} 
 & 6 & NA  & NA & 3339{*}  & \textbf{3342.50} & NA  & NA & 3420  & 3492.50 & 3343  & 3352.50\tabularnewline
\hline 
\end{tabular}

\label{Table10}

\end{table}

\subsection{Statistical Analysis}

Our statistical analysis of all the obtained results from Tables \ref{Table3} to \ref{Table10} is based on the $1 \times N$ pair comparisons. The Wilcoxon rank-sum test is used to assess whether the control strategy provides results that are significantly different from those of the other strategies.

To handle FWER errors due to multiple comparisons, we adopted the Bonferroni-Holm \cite{Ref45} correction for adjusting the value of $\alpha$ (based on $p_{holm} = \frac{\alpha}{i}$) in ascending order. For an i-ordered strategy, the $p-value$ $p_i$ is compared with the value of $p_{holm}$ for the same row of the table. In this study, $\alpha$ is set to 0.05 and 0.10 because most strategies are well-tuned and report their known best test suite sizes. Whenever the $p-value$ $p_i$ is less than the corresponding value of $p_{holm}$, the results imply that the test suite is smaller for the QLSCA than for the i-ordered strategy, which means that the QLSCA has a smaller median population. Table \ref{Table11} summarizes the overall statistical analysis.

\begin{table}

\caption{Wilcoxon Rank-Sum Tests for Tables \ref{Table3} till \ref{Table10} with QLSCA as control strategy}

\begin{center}
\includegraphics[width=0.99\linewidth]{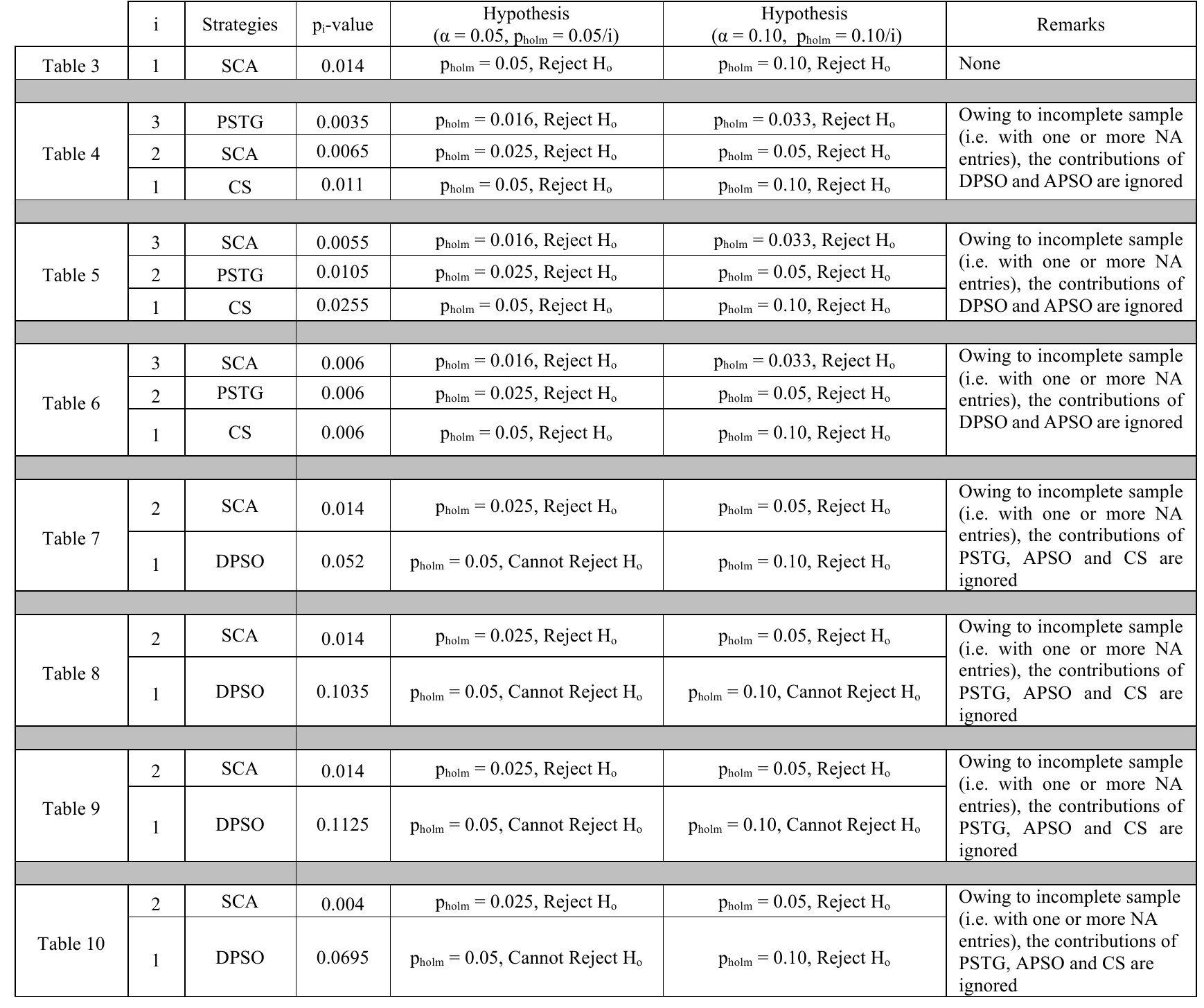}
\end{center}

\label{Table11}

\end{table}

\section{Experimental Observation}

Reflecting on the experimental results yields a number of observations.

Concerning the first set of experiments described in Section 7.1, Table \ref{Table3} compares the performances of the QLSCA and the SCA in terms of test size and execution time. We note that for all mean test sizes, the QLSCA outperforms the SCA (6 of 6). A similar trend can be seen for the best test size. The QLSCA outperforms the SCA in all cases. On a positive note, the SCA can match the result of the QLSCA for the $CA (N; 3, 4^6)$. As expected, the SCA outperforms the QLSCA in terms of execution time in all cases due to the overhead introduced by the Q-learning algorithm. Arguably, the trade-off between execution time and test size is necessary to ensure efficient, quality tests and to promote cost savings.

The box plot analyses of Table \ref{Table3} shown in Fig \ref{Figure8-BoxPlot} ($a$)--($f$) reveal a number of salient characteristics of the QLSCA and the SCA search processes. Considering the $CA (N; 2, 3^{13})$, even though they have the same quartile bias range (i.e., they incline towards the lower quartile) and the same top-to-bottom whisker width, the QLSCA has a lower median than the SCA. For the $CA (N; 2, 10^5)$, the box plot is symmetric in the case of the SCA. Unlike that of the SCA, when the outlying point is removed, the box plot for the QLSCA is biased towards the upper quartile (but with a lower top and bottom whisker width). Nevertheless, the median of the QLSCA is far lower than that of the SCA. In the $CA (N; 3, 4^6)$, the box plot for the SCA is asymmetric. The SCA also has a larger interquartile range and a higher mean than the QLSCA. Additionally, unlike the SCA, the QLSCA’s whisker width is also 0, indicating consistently better results for the 20 runs, except for one outlier. As far as the $MCA (N; 2, 5^1 3^8 2^2)$ is concerned, both the SCA and the QLSCA have the same bias towards the lower quartile. However, the QLSCA has a better median and top–to-bottom whisker width. As for the $MCA (N, 2, 6^1 5^1 4^6 3^8 2^3)$, both the SCA and the QLSCA have the same bias towards the upper quartile. Ignoring outliers, the interquartile range is smaller for the SCA than the QLSA. However, the median of the SCA is far greater than that of the QLSCA. Finally, in the case of the $MCA (N, 2, 7^1 6^1 5^1 4^6 3^8 2^3)$, the box plot for the SCA can be seen as a shifted version of the box plot of the QLSA (from left to right). Therefore, the median of the QLSCA is lower than that of the SCA.

Complementing the box plots, the convergence pattern analyses for the best run in Fig \ref{Figure9-convergence-pattern} ($a$) to ($f$) describe the convergence behaviour of the QLSCA and the SCA. With the exception of the $CA (N; 2, 10^5)$ and the $CA (N; 3, 4^6)$, all other cases ($CA (N; 2, 3^{13}$, $MCA (N; 2, 5^1 3^8 2^2)$, $MCA (N, 2, 6^1 5^1 4^6 3^8 2^3)$, and $MCA (N, 2, 7^1 6^1 5^1 4^6 3^8 2^3)$) indicate that the QLSCA converges faster than the SCA. Therefore, it has the potential to yield a smaller test suite size.

We note that the average search operator percentage distribution shown in Fig \ref{Figure10-Bar-Figure} is almost the same for all the search operations (nearly 25\%) involving uniform CAs ($CA(N; 2, 3^{13})$, $CA(N; 2, 10^5)$, and $CA(N; 3, 4^6)$). However, for non-uniform CAs ($MCA(N; 2, 5^1 3^8 2^2)$, $MCA(N, 2, 6^1 5^1 4^6 3^8 2^3)$, and $MCA(N, 2, 7^1 6^1 5^1 4^6 3^8 2^3)$), there is a clear tendency to favour crossover operation (i.e., with the highest average percentage for all 3 MCAs) and less tendency to employ Lévy flight motion (i.e., with the lowest average percentage for all 3 MCAs).

In the second set of experiments, the benchmark results highlight the overall performance of the QLSCA and the SCA in comparison with other meta-heuristic based strategies. Table \ref{Table4} demonstrates that the QLSCA outperforms all other strategies with respect to the best mean test size with 90\% (9 of 10 entries). DPSO and APSO produce the second-best with 20\% (2 of 10 entries) of the best mean test sizes. The PSTG, the CS, and the SCA do not contribute any of the best means. Concerning the best test size, the QLSCA also outperforms other strategies with 90\% (9 of 10 entries). Other strategies contribute 50\% of the best results (5 of 10 entries), except for the PSTG, which contributes only 30\% (3 of 10 entries).

In Table \ref{Table5}, we observe that the QLSCA has the best mean test size 77.7\% of the time (7 of 9 entries). The runner-up is the CS with 22.22\% (2 of 9 entries). The SCA and other strategies do not contribute to the best mean test size. Concerning the best test size, the QLSCA has the best performance with 66.66\% (6 of 9 entries). DPSO comes in second with 55.55\% (5 of 9 entries). APSO comes in third with 33.33\% (3 of 9 entries). The CS comes in fourth with 22.22\% (2 of 9 entries). Finally, the SCA and the PSTG come in last with 11.11\% (1 of 9 entries).

Concerning the results in Table \ref{Table6}, there is no contribution from other strategies because the QLSCA dominates the best mean test size with 100\% (8 of 8). As for the best test size, the QLSCA contributes 87.5\% (7 of 8 entries) and DPSO contributes 25\% (2 of 8 entries). APSO and the SCA contribute 12.5\% (1 of 8 entries). The CS and the PSTG perform the worst with no examples having the best test size.

As for Table \ref{Table7}, the QLSCA contributes 66.66\% (4 of 6 entries) as far as the best mean test size is concerned. The other best mean test sizes are shared by APSO and DPSO with 16.66\% (1 of 6 entries). DPSO performs the best performance with respect to the best test size with 66.66\% (4 of 6 entries). The QLSCA comes in second with 50\% (3 of 6 entries). APSO and the CS come in third with 16.66\% (1 of 6 entries). The PSTG and the SCA do not contribute as far as the best test size is concerned.

In Table \ref{Table8}, the QLSCA outperforms other strategies with 66.66\% (4 of 6 entries) as far as the mean test size is concerned. The PSTG and DPSO share the rest of the best mean test sizes with 16.66\% (1 of 6 entries). The other strategies do not contribute towards the best mean test size. A similar observation can be seen as far as the best test size is concerned. The QLSCA outperforms all other strategies with 66.66\% (4 of 6 entries). All the other strategies contribute at least 16.66\% (1 of 6 entries), except for the PSTG, which contributes 0\%.

According to Table \ref{Table9}, DPSO dominates as far as the best mean test size is concerned with 66.66\% (4 of 6 entries). The QLSCA and the CS come in second with 16.66\% (1 of 6 entries). The SCA and the other strategies do not contribute as far as the best mean test size is concerned. A similar observation can be made in the case of the best test size. DPSO outperforms all the other strategies with 66.66\% (4 of 6 entries). The QLSCA and the CS are second and contribute 16.66\% (1 of 6 entries). The SCA and the other strategies do not contribute to the best test size.

Concerning the results in Table \ref{Table10}, we observe that the QLSCA outperforms the other strategies with 77.77\% (7 of 9 entries) as far as the best mean test size is concerned. DPSO comes in second with 22.22\% (2 of 9 entries). Aside from the QLSCA and DPSO, no other strategies contribute towards the best mean test size. Similarly, no strategies contribute towards the best test size apart from the QLSCA and DPSO with 88.88\% (8 of 9 entries) and 44.44\% (4 of 9 entries), respectively.

Statistical analyses for Tables \ref{Table3} through \ref{Table10} (given in Table \ref{Table11}), indicate that the QLSCA statistically dominates all the state-of-the-art strategies at the 90\% confidence level. From the statistical analysis for Table \ref{Table3}, the significance of the QLSCA in comparison to the SCA is evident. The analyses for Tables \ref{Table4} through \ref{Table6} also support the alternate hypothesis (that the QLSCA performs better than the PSTG, the SCA, and the CS). The contributions of DPSO and APSO are ignored because results are unavailable for some CAs. Referring to the analyses for Tables \ref{Table7} to \ref{Table9}, the QLSCA is better than the SCA but not DPSO at the 95\% confidence level (excluding contributions from the PSTG, APSO, and the CS). However, at the 90\% confidence level, the QLSCA is better than DPSO for Tables \ref{Table7} and Table \ref{Table10}.

\section{Threats to Validity}

Normally, most of the research in this field addresses different threats during experiments and evaluations. These threats are to internal and external validity and depend on the type of research. This study is not infallible with respect to these threats. Threats to external validity occur when we cannot generalize experiments to real-world problems. Here, there is no guarantee that the adopted benchmarks represent real-world applications with the same number of parameters and values and the same interaction strength. We have tried to eliminate this threat by choosing the most common and realistic benchmarks in the literature for the experiments. These benchmarks are widely used for evaluations and have been selected from real configurable software or obtained from a simulation of possible configurations.

Threats to internal validity are concerned with the factors that affect the experiments without our knowledge and/or are out of our control. The differences in population size, the number of iterations and parameter settings of each meta-heuristic based strategy are examples of threats to internal validity. Because source code is not available for all implementations, we cannot ensure that the compared strategies have the same number of fitness function evaluations as the QLSCA. Despite these differences, we believe that our comparison is valid because the published test size results are obtained using the best control parameter settings and are not affected by the operating environment. In fact, in addition to the best size results, we relied on the mean results to ascertain the performance of each strategy due to the randomness of each meta-heuristic run.

Another threat to internal validity is the generation time for each strategy. It is well-known that the size of the test suite is not affected by the environment. However, the generation time for the test suite is strongly affected by the running environment. Therefore, we cannot directly compare the generation time with published results. Indeed, to compare the generation time fairly, it is necessary that all strategies be implemented and used in the same environment. In fact, in many cases, the strategies may need to be implemented in the same programming language using the same data structure (in addition to from running for the same number of iterations).

Finally, the choice of unsupervised reinforcement learning based on the Q-learning algorithm may be another threat to internal validity. State-action-reward-state-action (SARSA) \cite{Ref31}, a competitor to Q-learning, could also be chosen for the QLSCA. Unlike Q-learning, which exploits look-ahead rewards, SARSA obtains rewards directly from the actual next state. We believe that because most of the time, the look-ahead reward eventually becomes the actual reward (except when there is a tie in the Q-table), the choice between SARSA and Q-learning is immaterial and results in no significant difference in performance.

\section{Concluding Remarks}

In this paper, we have described a novel hybrid QLSCA that uses a combination of the sine, cosine, Lévy flight, and crossover operators. Additionally, we have applied the QLSCA to the combinatorial test suite minimization problem as our benchmark case study.

The intertwined relationship between exploration and exploitation in both Q-learning and the QLSCA strategy needs to be highlighted. As far as the Q-learning algorithm is concerned, exploration and exploitation deal with online updating of (learned) Q-table values to identify promising search operators for future selection (using rewards and punishments). Initially, Q-learning favours exploration, but in later iterations, it favours exploitation (using a probabilistic value that decreases over time). Unlike Q-learning, the QLSCA’s exploration and exploitation obtain the best possible solutions by dynamically executing the right search operator at the right time. Specifically, the exploration and exploitation of the QLSCA work synergistically with the Q-table. With the help of the Q-table, the QLSCA can eliminate the switch parameter $r_4$ defined in the SCA (refer to Eq. 1 and Eq. 2). Therefore, the QLSCA, unlike the SCA, can adaptively identify the best operation based on the learned Q-table values. In this manner, the QLSCA’s decision to explore or exploit (i.e., choosing the best search operator at any point in the search process) is directly controlled by the learned Q-table values.

Concerning the ensemble of operators, the introduction of crossover and Lévy flight within the QLSCA helps enhance the solution diversity and provides a mechanism for leaving local extrema. In addition to the fixed switching probability and the bounded magnitude of the sine and cosine functions, the fact that the sine and cosine operators are mathematically related (see Eq. 14 and Eq. 15) can be problematic.

\begin{equation}
cos\theta=sin(\frac{\pi}{2}-\theta)
\end{equation}

\begin{equation}
sin\theta=cos(\frac{\pi}{2}-\theta)
\end{equation}

As far as intensification and diversification are concerned, the use of either sine or cosine may cause the search process to become stuck at a local minimum (because they alternate from -1 to 1). Consequently, the performance of the SCA appears to be poorer than that of the QLSCA (and the other strategies) in almost all cases. In fact, the box plot (see Fig \ref{Figure8-BoxPlot}) and the convergence pattern analyses (see Fig \ref{Figure9-convergence-pattern}) confirm our observation. On a positive note, the SCA runs much faster than the QLSCA. The introduction of additional operators and Q-learning cuts the execution time in half compared to the original SCA.

Considering the average search operator percentage distribution, we observe the following patterns based on our experiments.

\begin{itemize}

\item For uniform CAs, the search operators are almost equally distributed. In such a situation, the Q-learning mechanism gives each search operation an equal opportunity to undertake the search.

\item For non-uniform CAs (MCAs), Q-learning is more inclined towards the crossover operation. Unlike the uniform CAs, the MCAs depend on number of parameter matchings for each test case in the test suite being different. Its ability to flexibly serve for both local and global searches is perhaps the main reason for Q-learning to favour the crossover operation. However, Lévy flight is less preferred by Q-learning because the resulting values are often too extreme and cause out-of-boundary parameter matching. When reflected back inside the boundary, the selected parameter is always reset to the boundary of the other endpoint (as an absorbing wall), which inadvertently promotes less diverse solutions.

\end{itemize}

In terms of the overall performance, in addition to surpassing the original SCA, the QLSCA has also outperformed many existing strategies by offering the best means in most of the table cell entries (the closest competitor is DPSO). Our statistical analyses support this observation. Putting DPSO aside, when $\alpha=0.05$, the QLSCA statistically outperforms the original SCA, the PSTG, APSO and the CS in all configurations given in Tables \ref{Table3} to \ref{Table10}. When $\alpha=0.10$, the QLSCA is statistically better than DPSO in two of four configuration tables (Tables \ref{Table7} and \ref{Table10}). Therefore, we believe that the QLSCA offers another useful alternative strategy for solving the $t-way$ test suite minimization problem.

The scope of future work includes our current evaluation of applying the QLSCA to other well-known optimization problems (e.g., timetabling and vehicle-routing problems) because of its performance. Additionally, we are investigating the comparative performance of the case-based reasoning approach and fuzzy inference systems with the Q-learning approach for the QLSCA.

\section*{Acknowledgments}
The work reported in this paper is funded by Fundamental Research Grant from Ministry of Higher Education Malaysia titled: A Reinforcement Learning Sine Cosine based Strategy for Combinatorial Test Suite Generation. We thank MOHE for the contribution and support, Grant number: RDU170103.


\end{document}